\documentclass[sigconf]{acmart}

\AtBeginDocument{%
  }

\copyrightyear{2024}
\acmYear{2024}
\setcopyright{rightsretained}
\acmConference[WSDM '24]{Proceedings of the 17th ACM International
Conference on Web Search and Data Mining}{March 4--8, 2024}{Merida, Mexico}
\acmBooktitle{Proceedings of the 17th ACM International Conference on Web
Search and Data Mining (WSDM '24), March 4--8, 2024, Merida,
Mexico}\acmDOI{10.1145/3616855.3635788}
\acmISBN{979-8-4007-0371-3/24/03}

\usepackage{xspace}
\usepackage{graphicx}
\usepackage{subfigure}
\usepackage{adjustbox}
\usepackage{algorithm}
\usepackage{algpseudocode}
\usepackage{wrapfig}
\usepackage{caption}
\usepackage{lipsum}
\usepackage{booktabs}
\usepackage{subfigure}
\usepackage{hyperref}
\usepackage{url}
\usepackage{array}

\newcommand{\name}[0]{DeSCo\xspace}
\newcommand{\xhdr}[1]{{\noindent\bfseries #1}.}
\newcommand{\rom}[1]{\textup{\uppercase\expandafter{\romannumeral#1}}}
\newcommand{\circleNum}[1]{\textcircled{\raisebox{-0.9pt}{#1}}}

\newtheorem{theorem}{Theorem}
\newtheorem{corollary}{Corollary}[theorem] 
\newtheorem{definition}{Definition}[section]
\newtheorem{lemma}{Lemma}[section]

\newcommand{\revision}[1]{#1}

\begin{document}

\title{DeSCo: Towards Generalizable and Scalable Deep Subgraph Counting}

\author{Tianyu Fu}
\affiliation{%
  \institution{Tsinghua University}
  \streetaddress{30 Shuangqing Rd}
  \state{Beijing}
  \country{China}}
\email{fty22@mails.tsinghua.edu.cn}

\author{Chiyue Wei}
\affiliation{%
  \institution{Tsinghua University}
  \streetaddress{30 Shuangqing Rd}
  \state{Beijing}
  \country{China}}
\email{chiyue.wei@duke.edu}

\author{Yu Wang}
\authornote{Corresponding Author}
\affiliation{%
  \institution{Tsinghua University}
  \streetaddress{30 Shuangqing Rd}
  \state{Beijing}
  \country{China}}
\email{yu-wang@tsinghua.edu.cn}

\author{Rex Ying}
\affiliation{%
  \institution{Yale University}
  \city{New Haven}
  \state{Connecticut}
  \country{USA}}
\email{rex.ying@yale.edu}

\renewcommand{\shortauthors}{Fu et al.}

\begin{abstract}
Subgraph counting is the problem of counting the occurrences of a given query graph in a large target graph. 
Large-scale subgraph counting is useful in various domains, such as motif analysis for social network and loop counting for money laundering detection.
Recently, to address the exponential runtime complexity of scalable subgraph counting, neural methods are proposed.
However, existing approaches fall short in three aspects.
Firstly, the subgraph counts vary from zero to millions for different graphs, posing a much larger challenge than regular graph regression tasks.
Secondly, current scalable graph neural networks have limited expressive power and fail to efficiently distinguish graphs for count prediction.
Furthermore, existing neural approaches cannot predict query occurrence positions.

We introduce \name, a scalable neural deep subgraph counting pipeline, designed to accurately predict both the count and occurrence position of queries on target graphs post single training.
Firstly, \name uses a novel \emph{canonical partition} and divides the large target graph into small neighborhood graphs, greatly reducing the count variation while guaranteeing no missing or double-counting.
Secondly, \emph{neighborhood counting} uses an expressive subgraph-based heterogeneous graph neural network to accurately count in each neighborhood. 
Finally, \emph{gossip propagation} propagates neighborhood counts with learnable gates to harness the inductive biases of motif counts. 
\name is evaluated on eight real-world datasets from various domains. It outperforms state-of-the-art neural methods with 137$\times$ improvement in the mean squared error of count prediction, while maintaining the polynomial runtime complexity.
Our open-source project is at \url{https://github.com/fuvty/DeSCo}.

\end{abstract}

\ccsdesc[500]{Information systems~Graph-based database models}

\keywords{subgraph counting, graph mining, graph neural network}

\received{10 August 2023}
\received[accepted]{10 October 2023}

\maketitle

\section{Introduction}\label{sec:intro}


Given a \emph{query} graph and a \emph{target} graph, the problem of subgraph counting is to count the number of \emph{patterns}, defined as subgraphs of the target graph, that are graph-isomorphic to the query graph~\cite{ribeiro2021survey}.

Subgraph counting is crucial for domains including biology~\cite{takigawa2013graph,sole2008spontaneous,adamcsek2006cfinder,bascompte2005simple,bader2003an}, social science~\cite{uddin2013dyad,prell2008looking,kalish2006psychological,wasserman1994social}, risk management~\cite{ribeiro2017shaping, akoglu2013anomaly}, and software analysis~\cite{valverde2005network,wu2018software}. 
For example, in brain networks, it is used to identify important functional motifs and understand how the brain evolves~\cite{sporns2004motifs}. In social networks, counts of stars, holes, or paths are used to characterize circles of friends~\cite{charbey2019stars}.

While being essential in graph and network analysis, subgraph counting is a \#P-complete problem~\cite{valiant1979complexity}. 
Due to the computational complexity, existing exact counting algorithms are restricted to small query graphs with no more than 5 vertices~\cite{pinar2017escape, ortmann2017efficient, ahmed2015efficient}. The commonly used VF2~\cite{cordella2004sub} algorithm fails to even count a single query of a 5-node chain within a week's time budget on a large target graph Astro~\cite{leskovec2007graph} with nineteen thousand nodes. 


\begin{figure}[tb]
    \centering
    \includegraphics[width=0.45\textwidth]{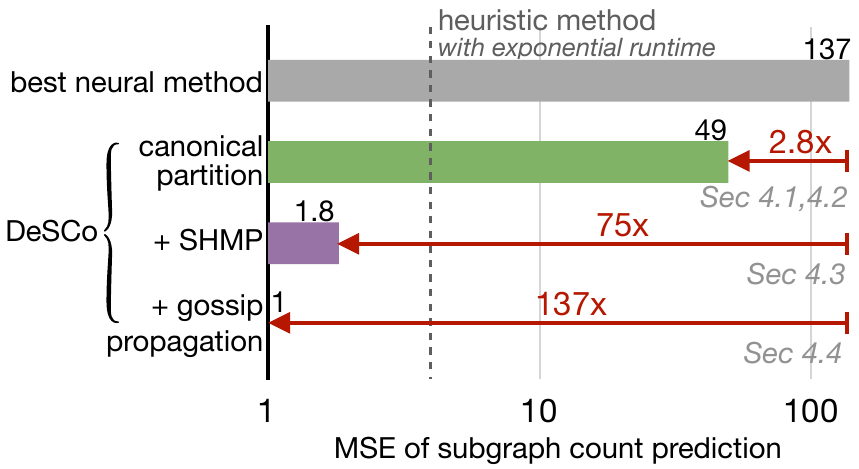}
    \vspace{-6pt}
    \caption{\name Pipeline reduces the mean square error (MSE) of subgraph count prediction with three components: canonical partition, subgraph-based heterogeneous message passing (SHMP) and gossip propagation. The MSE is evaluated and averaged on eight real-world datasets.}
    \label{fig:intro/benefit}
    \vspace{-10pt}
\end{figure}

Luckily, approximate counting of query graphs is sufficient in most real-world use cases~\cite{iyer2018asap, kashtan2004efficient, ribeiro2010efficient}. 
Heuristic methods can scale to large targets by substructure sampling, random walk, and color-based sampling, allowing estimation of the frequency of query graph occurrences. However, they still cannot scale to large queries.
Very recently, Graph Neural Networks (GNNs) are employed as a deep learning-based approach to scale the query graphs in subgraph counting~\cite{Zhao2021ALS, Liu2020NeuralSI, Chen2020CanGN}. The target graph and the query graph are embedded via a GNN, which predicts the motif count through a regression task.

\begin{figure}[tb]
    \centering
    \includegraphics[width=0.45\textwidth]{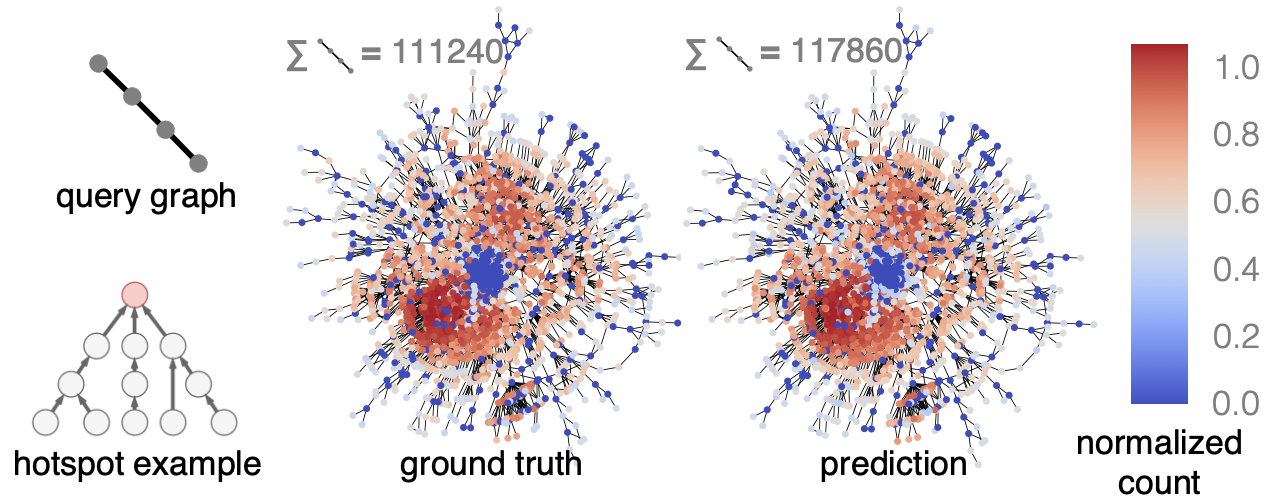}
    \vspace{-6pt}
    \caption{
        The total count and the position distribution of the query graph over the CiteSeer Citation Network. 
        The figure compares between ground truth and \name predictions.
        The hotspots are where the 4-chain patterns appear most often in CiteSeer.
    }
    \label{fig:distribution}
    \vspace{-10pt}
\end{figure}

However, there exist several major challenges with existing heuristic and GNN approaches:
1) The number of graph structures and count variation both grow super-exponentially with respect to the graph size~\cite{sloane2014handbook, read1998atlas}, resulting in large approximation error~\cite{ribeiro2021survey}. For different large target graphs, the counts of the same query can vary from zero to millions, making the task much harder than most graph regression tasks~\cite{sorkun2019aqsoldb}, which only predict a single-digit number with a small upper bound.
2) The expressive power of commonly used message passing GNNs is limited by the Weisfeiler-Lehman (WL) test~\cite{leman1968reduction, Chen2020CanGN, xu2018powerful}. Certain structures are not distinguishable with these GNNs, let alone counting them, resulting in the same count prediction for different queries.
3) Furthermore, most existing approximate heuristic and GNN methods only focus on estimating the total count of a query in the target graph~\cite{bressan2019motivo, Liu2020NeuralSI, Chen2020CanGN, bressan2021faster}, but not the occurrence positions of the patterns, as shown in Figure~\ref{fig:distribution}.
Yet such position distribution information is crucial in various applications~\cite{yin2019local, tsourakakis2017scalable, benson2016higher,faust2010puzzle, holland1976local}.

\xhdr{Proposed work}
To resolve the above challenges, we propose \name, a GNN-based model that learns to predict both pattern counts and occurrence positions on any target graph.
The main idea of \name is to leverage the local information of neighborhood patterns to predict query counts and occurrences in the entire target graph. \name first uses \emph{canonical partition} to decompose the target graph into small neighborhoods.
The local information is then encoded using a GNN with \emph{subgraph-based heterogeneous message passing}.
Finally, we perform \emph{gossip propagation} to use inductive biases to improve counting accuracy over the entire graph.
Our contributions are four-fold.


\xhdr{Canonical partition}
Firstly, we propose \emph{canonical partition} that divides the problem into subgraph counting for individual neighborhoods. 
We theoretically prove that no pattern will be double counted or missed for all neighborhoods.
The algorithm allows the model to make accurate predictions on large target graphs with high count variation.
Furthermore, we can predict the pattern position distribution for the first time, as shown in Figure~\ref{fig:distribution}. 
In this citation network, the hotspots represent overlapped linear citation chains, indicating original publications that motivate multiple future directions of incremental contributions~\cite{gao2017testing, yang2015using}, which shed light on the research impact of works in this network.

\xhdr{Subgraph-based heterogeneous message passing}
Secondly, we propose a general approach to enhance the expressive power of any MPGNNs by encoding the subgraph structure through heterogeneous message passing. The message type is determined by whether the edge presents in a certain subgraph, e.g., a triangle.
We show that this architecture outperforms expressive GNNs, including GIN~\cite{xu2018powerful} and ID-GNN~\cite{you2021identity}, while maintaining the polynomial runtime complexity for scalable subgraph counting.

\xhdr{Gossip propagation}
We further improve the count prediction accuracy by utilizing two inductive biases of the counting problem: homophily and antisymmetry. 
Real-world graphs share similar patterns among adjacent nodes, as shown in Figure~\ref{fig:distribution}.
Furthermore, since canonical count depends on node indices, there exists antisymmetry due to canonical partition.
Therefore, we propose a \emph{gossip propagation} phase featuring a learnable gate for propagation to leverage the inductive biases.

\xhdr{Generalization Framework}
We propose a generalization framework that uses the carefully designed synthetic dataset to enable model generalization to different real-world datasets. After training on the synthetic dataset, the model can directly perform subgraph counting inference with high accuracy on real-world datasets.

To demonstrate the effectiveness of \name, we compare it against state-of-the-art GNN-based subgraph counting methods~\cite{Chen2020CanGN, Liu2020NeuralSI,liu2022dualMessage}, as well as approximate heuristic method~\cite{bressan2019motivo, bressan2021faster} on eight real-world datasets from various domains.
\name achieves $137\times$ mean square error reduction of count predictions for both small and large targets, as shown in Figure~\ref{fig:intro/benefit}. To the best of our knowledge, it is also the first approximate method to accurately predict pattern position distribution as illustrated in Figure~\ref{fig:distribution}.
\name also maintains polynomial runtime efficiency, demonstrating orders of magnitude speedup over the heuristic~\cite{bressan2019motivo, bressan2021faster} and exact methods~\cite{cordella2004sub, sun2020memory}.

\begin{figure*}[t]
    \centering
    \includegraphics[width=\textwidth]{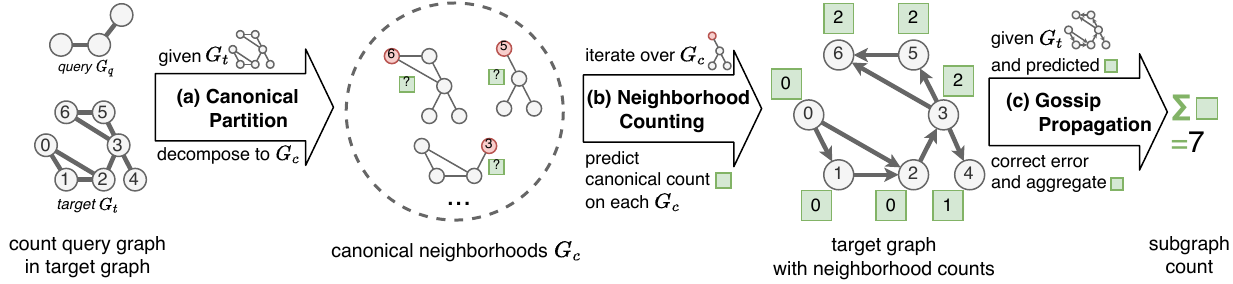}
    \vspace{-15pt}
    \caption{\name Pipeline in 3 steps. \textbf{(a) Step 1. Canonical Partition}: Given \emph{query} and \emph{target}, decompose \emph{target} into multiple node-induced subgraphs, i.e., \emph{canonical neighborhood}s, based on node indices. Each neighborhood contains a \emph{canonical node} that has the greatest index in the neighborhood. \textbf{(b) Step 2. Neighborhood Counting}: Predict the \emph{canonical count}s of each neighborhood via an expressive GNN, and assign the count of the neighborhood to the corresponding \emph{canonical node}. Neighborhood counting is the local count of queries. \textbf{(c) Step 3. Gossip Propagation}: Use GNN prediction results to estimate \emph{canonical count}s on the \emph{target} graph through learnable gates.
    }
    \label{fig:pipeline}
    \vspace{-5pt}
\end{figure*}

\section{Related Works}\label{sec:related_work}

There has been extensive lines of work for subgraph counting.

\xhdr{Exact counting algorithms} Exact methods generally count subgraphs by searching through all possible node combinations and finding the matching pattern. Early methods usually focus on improving the matching phase~\cite{wernicke2006fanmod, cordella2004sub, milo2002network}
Recent approaches emphasize the importance of pruning the search space and avoiding double counting~\cite{demeyer2013index, mawhirter2019graphzero, shi2020graphpi, mawhirter2019automine}, which inspires the design our canonical count objective (Section~\ref{sec:pipeline/canonical_count}).
However, exact methods still scale poorly in terms of query size (often no more than five nodes) despite much effort~\cite {pinar2017escape, chen2020dwarvesgraph}.

\xhdr{Approximate heuristic methods} To further scale up the counting problem, approximate counting algorithms sample from the target graph to estimate pattern counts. Strategies like path sampling~\cite{wang2017moss,jha2015path}, random walk~\cite{yang2018ssrw, saha2015finding}, substructure sampling~\cite{fu2020lessmine, iyer2018asap}, and color coding~\cite{bressan2018motif, bressan2019motivo, bressan2021faster} are used to narrow the sample space and provides better error bound. 
However, large and rare queries are still hard to find in the vast sample space, leading to large approximation error~\cite{bressan2019motivo, bressan2021faster}.

\xhdr{GNN-based approaches} Recently, GNNs have been used to attempt counting large queries. \cite{Liu2020NeuralSI,NEURIPS2018_e77dbaf6} use GNNs to embed the query and target graph, and predict subgraph counts via embeddings. \cite{Chen2020CanGN} theoretically analyzes the expressive power of GNNs for counting and proposes an expressive GNN architecture. \cite{Zhao2021ALS} proposes an active learning scheme for the problem. \cite{liu2022dualMessage} proposes expensive edge-to-vertex dual graph transformation to enhance the model expressive power for subgraph counting.
Unfortunately, large target graphs have extremely complex structures and a high variation of pattern count, so accurate prediction remains challenging.
\section{Preliminary}\label{sec:preliminary}

Let $G_t=(V_t,E_t)$ be a large \emph{target} graph with vertices $V_t$ and edges $E_t$. Let $G_q=(V_q,E_q)$ be the \emph{query} graph of interest. 
The \emph{subgraph counting problem} $\mathcal{C}(G_q,G_t)$ is to calculate the size of the \emph{set of patterns} $\mathcal{P}=\{G_p|G_p\subseteq G_t\}$ in the target graph $G_t$ that are isomorphic to the query graph $G_q$, that is, $\exists$ bijection $f:V_p \mapsto V_q$ such that $(f(v),f(u))\in E_q$ if and only if $(v,u)\in E_p$, denoted as $G_p \cong G_q$.

Subgraph counting can be categorized into induced and non-induced counting~\cite{ribeiro2021survey}.
A subgraph \( G_p = (V_p, E_p) \) of \( G_t \) is an induced subgraph if it satisfies two conditions: \( V_p \subseteq V_t \) and for any two vertices \( u, v \in V_p \), they are adjacent in \( G_p \) if and only if they are adjacent in \( G_t \). This relationship is denoted as $G_p \subseteq G_t$. 
Without loss of generality, we focus on the connected, induced subgraph counting problem, following modern mainstream graph processing frameworks~\cite{hagberg2008networkx,peixoto2014graphtool} and real-world applications~\cite{wong2012biological, milo2002network}. It is also possible to obtain non-induced occurrences from induced ones with a transformation~\cite{floderus2015induced}. 
Our GNN approach can natively support graphs with node features and edge directions. But in alignment with exact and heuristic methods, we use undirected graphs without node features in experiments to investigate the ability to capture graph topology.
\section{\name Pipeline}\label{sec:pipeline}

In this section, we introduce the pipeline of \name as shown in Figure~\ref{fig:pipeline}.
To perform subgraph counting, \name first performs \textbf{canonical partition} to decompose the target graph to many canonical neighborhood graphs. Then, \textbf{neighborhood counting} uses the subgraph-based heterogeneous GNN to embed the query and neighborhood graphs and performs a regression task to predict the canonical count on each neighborhood. Finally, \textbf{gossip propagation} propagates neighborhood count predictions over the target graph with learnable gates to further improve counting accuracy. 
We will first introduce the model objective before elaborating on each step.


\subsection{Canonical Count Objective}\label{sec:pipeline/canonical_count}

\begin{figure}[b]
    \vspace{-6pt}
    \centering
    \includegraphics[width=0.45\textwidth]{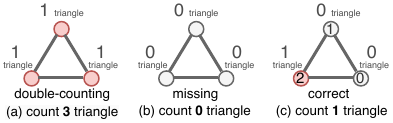}
    \vspace{-6pt}
    \caption{When counting, (a) double-counts and (b) misses the triangle in the neighborhoods due to symmetry. (c) \name uses the canonical node to break symmetry and correctly count the triangle. \circleNum{i} are the node indices.}
    \label{fig:canonical_count}
    \vspace{-8pt}
\end{figure}

\begin{figure*}[t]
    \vspace{-5pt}
    \centering
    \includegraphics[width=0.85\textwidth]{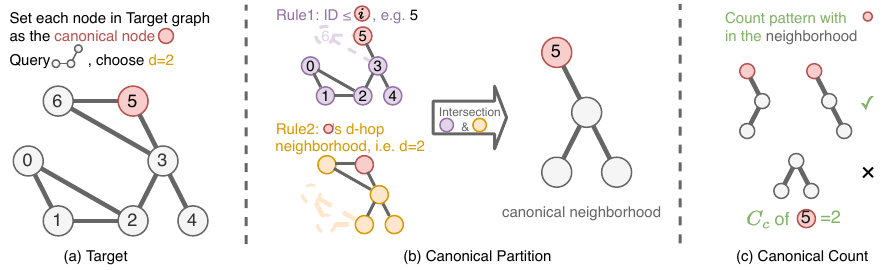}
    \vspace{-7pt}
    \caption{An example of canonical partition and canonical count. (a) Choose node 5 from the target graph as the \emph{canonical node} (red circle). (b) \emph{Canonical partition} generates the corresponding \emph{canonical neighborhood} graph. It performs an ID-restricted breadth-first search to find the induced neighborhood that complies with both Rule 1 and Rule 2. (c) The corresponding \emph{canonical count} is defined by the number of patterns containing the canonical node in the canonical neighborhood. \name's \emph{neighborhood counting} phase predicts the canonical count for each canonical neighborhood. 
    }
    \label{fig:canonical_objective}
    \vspace{-8pt}
\end{figure*}


\xhdr{Motivation}
For commonly seen node-level tasks such as node classification, each node is responsible for predicting its own node value. However, for subgraph counting, since each pattern contains multiple nodes, it is unclear which node should be responsible for predicting the pattern's occurrence.
As illustrated in Figure~\ref{fig:canonical_count}, the ambiguity can lead to missing or double-counting of the motif, especially for queries with symmetric nodes, e.g. triangle.
So we propose the canonical count objective to eliminate the ambiguity by assigning a specific canonical node responsible for each pattern.
The canonical node is used to represent the pattern position.
The canonical count is used as the local count prediction objective for the GNN and gossip propagation. 

To break the symmetry, we randomly assign node indices on the target graph and define the \emph{canonical node}.

\begin{definition}[canonical node]
    \emph{Canonical node} $v_c$ is the node with the largest node index in the pattern. 
    \begin{equation}
        v_c = \max_{I}V_p
    \end{equation}
    \label{def:canonical_node}
\end{definition}

Based on the index, we assign the count of the k-node pattern to its \emph{canonical node} and define \emph{canonical count}.

\begin{definition}[canonical count]
    \emph{Canonical count} $\mathcal{C}_c$ equals the number of patterns that share the same canonical node.
    \begin{equation}
        \mathcal{C}_c(G_q, G_t, v_c) = |\{G_p \subseteq G_t | G_p  \cong G_q, v_c=\max_{I}V_p \}|
        \label{eq:canonical_count}
    \end{equation}
    \label{def:canonical_count}
\end{definition}

The canonical count $\mathcal{C}_c(G_q, G_t, v_c)$ differs from the regular count $\mathcal{C}_c$, as it takes an additional variable - a node $v_c$ from the target graph.
As shown in Figure~\ref{fig:canonical_count}(c), a pattern is only counted by its canonical node in $\mathcal{C}_c$. 
So the summation of $\mathcal{C}_c$ over all nodes equals the count of all patterns, $\mathcal{C}$, as stated in Lemma~\ref{lm:count_canonical_count} and proven in Appendix~\ref{sec:appendix/prof_lemma4d1}.

\begin{lemma}
    The subgraph count $\mathcal{C}$ of query in target equals the summation of the canonical count of query in target for all target nodes.

    \begin{equation}
        \mathcal{C}(G_q,G_t)=\sum_{v_c \in V_t} \mathcal{C}_c(G_q,G_t,v_c)
        \label{eq:count_canonical_count}
    \end{equation}

    \label{lm:count_canonical_count}
\end{lemma}

\xhdr{Advantage} By predicting the canonical count of each node, \name can naturally get the pattern position distribution.

Lemma~\ref{lm:count_canonical_count} allows the decomposition of the counting problem into multiple canonical count objectives. We use the following canonical partition to minimize the overhead for the decomposition. 


\subsection{Canonical Partition}

\xhdr{Motivation} 
In Lemma~\ref{lm:count_canonical_count}, each canonical count $\mathcal{C}_c$ is obtained with the entire target graph $G_t$. 
In order to overcome the high computational complexity, we partition the target to reduce the graph size for the canonical count.
We observe that each canonical count only depends on some local neighborhood structure as shown in Figure~\ref{fig:canonical_objective}(c).
So we propose \emph{canonical partition} to efficiently get the small neighborhood.

\xhdr{Unique challenges of partition for canonical count} 
Commonly used graph partition strategies include cutting edges~\cite{bader2013graph} and taking d-hop neighborhoods~\cite{hamilton2017inductive}.
However, edge-cutting breaks the pattern structure, leading to incorrect count; D-hop neighborhoods guarantee correctness, yet are unnecessarily large since patterns exist in many overlapping neighborhoods.

Thus, we define \emph{canonical partition}. It neglects the neighborhood structure that does not influence the canonical count of each node. Canonical partition uses node indices to filter nodes as illustrated in Figure~\ref{fig:canonical_objective}(a), (b).

\begin{definition}[canonical partition]
    \emph{Canonical partition} $\mathcal{P}$ crops the index-restricted d-hop neighborhood around the center node from the target graph. $\mathcal{D}(G_t,v_i,v_c)$ means the shortest distance between $v_i$ and $v_c$ on $G_t$.
    \begin{equation}
    \begin{aligned}
        &\mathcal{P}(G_t, v_c, d) = G_c,\\ 
        &\operatorname{ s.t. } G_c \subseteq G_t, V_c = \{ v_i \in V_t|\mathcal{D}(G_t,v_i,v_c) \leq d , v_i \leq v_c\}
        \label{eq:canonical_neighborhood}
    \end{aligned}
    \end{equation}
    \label{def:canonical_partition}
\end{definition}

The graph $G_c$ obtained by canonical partition is called the \emph{canonical neighborhood}. Canonical neighborhoods can correctly substitute the target graph in canonical count as proven in Appendix~\ref{sec:appendix/theo1}. Thus, we derive Theorem~\ref{thm:count_partition_count}.

\begin{theorem}
    The subgraph count of query in target equals the summation of the canonical count of query in canonical neighborhoods for all target nodes. Canonical neighborhoods are acquired with canonical partition $\mathcal{P}$, given any $d$ greater than the diameter of the query.
    \begin{equation}
    \begin{aligned}
        \mathcal{C}(G_q,G_t) &= \sum_{v_c \in V_t} \mathcal{C}_c(G_q,\mathcal{P}(G_t, v_c, d),v_c), \\
        d &\geq \max_{v_i, v_j \in V_q}\mathcal{D}(G_q, v_i, v_j)
        \label{eq:count_partition_count}
    \end{aligned}
    \end{equation}
    \label{thm:count_partition_count}
\end{theorem}

In \name, given the target graph $G_t$, it iterates over all nodes $v$ of the target $G_t$ and divides it into a set of canonical neighborhoods $G_{v_c}$ with \textbf{canonical partition}.
In practice, we set $d$ as the maximum diameter of query graphs to meet the requirements of Theorem.\ref{thm:count_partition_count}. 
See Appendix~\ref{sec:appendix/implement_canonical_partition} for the implementation of $\mathcal{P}(G_t, v_c, d)$.

\xhdr{Advantage} Canonical partition dramatically reduces the worst and average complexity of the subgraph counting problem by a factor of $1/10^{70}$ and $1/10^{11}$, thanks to the sparse nature of real-world graphs (discussed in Appendix~\ref{sec:appendix/complexity_benefit}). 
Furthermore, diverse target graphs can have similar and limited kinds of canonical neighborhoods. So it boosts the generalization power of \name as shown in Section~\ref{sec:generalization}.

This divide-and-conquer scheme not only greatly reduces the complexity of each GNN prediction, but also makes it possible to predict the count distribution over the entire graph.
After the canonical partition, \name uses the following model to predict the canonical count for each decomposed neighborhood.

\subsection{Neighborhood Counting}

After canonical partition, GNNs are used to predict the \emph{canonical count} $C_c(G_q, G_{v_c}, v_c)$ on any canonical neighborhood $G_{v_c}$ in the \textbf{neighborhood counting} stage. 
The canonical neighborhood and the query are separately embedded using GNNs. The embeddings are passed to a multilayer perceptron to predict the canonical count. 





\xhdr{Motivation}
Previous work~\cite{Chen2020CanGN} shows message passing (MP) GNNs confuse certain graph structures and harm the counting accuracy. 
To enhance GNN's expressive power while remaining scalable, we propose the Subgraph-based Heterogeneous Message Passing (SHMP) framework. Inspired by~\cite{Morris2019WeisfeilerAL}, SHMP incorporates subgraph information to boost the expressive power. 
In the meantime, SHMP avoids using super-node~\cite{Morris2019WeisfeilerAL} or message permutation~\cite{Chen2020CanGN} that are computationally expensive during message passing. 

\begin{figure}[tb]
    \centering
    \includegraphics[width=0.37\textwidth]{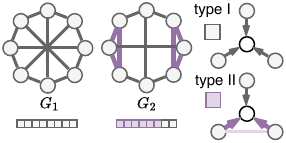}
    \vspace{-6pt}
    \caption{
        Proposed SHMP. Embedded with regular MP, graphs $G_1$ and $G_2$ are indistinguishable. While embedded with SHMP, $G_2$ is successfully distinguished with six type \rom{2} node embeddings, demonstrating better expressive power of SHMP.
    }
    \label{fig:SHMP}
    \vspace{-8pt}
\end{figure}        

\xhdr{Neighborhood counting with SHMP} To embed the input graph, SHMP uses small subgraph structures to categorize edges into different edge types, and uses different learnable weights for each edge type. 

\begin{definition}[subgraph-based heterogeneous message passing]
    The SHMP computes each node's representation with equation~\ref{eq:SHMP}. Here $k$ denotes the layer; $\gamma$ denotes the update function; $\phi_h^k$ denotes the message function of the h-th edge type; $N_h(i)$ denotes nodes that connect to node i with the h-th edge type; $\textsc{Agg}$ and $\textsc{Agg}'$ are the permutation invariant aggregation function such as summation. 
    
    \begin{equation}
    \begin{aligned}
        \mathbf{x}_{i}^{(k)} &= \gamma^{(k)}\bigg(\mathbf{x}_{i}^{(k-1)},\textsc{Agg}'_{h \in H} ( M_h ) \bigg) \\
        M_h &=  \textsc{Agg}_{j \in N_h(i)} \big( \phi^{(k)}_h(\mathbf{x}_{i}^{(k-1)}, \mathbf{x}_{j}^{(k-1)}, \mathbf{e}_{j, i} ) \big)
        \label{eq:SHMP}
    \end{aligned}
    \end{equation}

\end{definition}
Note that MP defined by major GNN frameworks~\cite{Fey2019pyg, wang2019dgl} is just a special case of SHMP if only one edge type is derived with the subgraph structure. We prove that SHMP can exceed the upper bound of MP in terms of expressiveness in Appendix~\ref{sec:appendix/prof_expressive_power}. 

For example, Figure~\ref{fig:SHMP} demonstrates that triangle-based heterogeneous message passing has better expressive power. Regular MPGNNs fail to distinguish different d-regular graphs $G_1$ and $G_2$ because of their identical type \rom{1} messages and embeddings, which is a common problem of MPGNNs~\cite{you2021identity}.
SHMP, however, can discriminate the two graphs by giving different embeddings. The edges are first categorized into two edge types based on whether they exist in any triangles (edges are colored purple if they exist in any triangles). Since no triangles exist in $G_2$, all of its nodes still receive type \rom{1} messages. While some nodes of $G_1$ now receive type \rom{2} messages with two purple messages and one gray message in each layer. As a result, the model acquires not only the adjacency information between the message sender and receiver, but also information among their neighbors. Such subgraph structural information improves expressiveness by incorporating high-order information in both the query and the target.
In \name, the canonical node of the neighborhood is also treated as a special node type in the heterogeneous message passing. 

\xhdr{Advantage} The triangle-based SHMP reduces the typical error of MPGNNs by 68\% as discussed in Appendix~\ref{sec:appendix/SHMP/regular_graph}, while remaining polynomial runtime complexity of $O(V+E^{3/2})$ as discussed in Appendix~\ref{sec:appendix/runtime}. The comparison with other expressive GNNs are shown in Table~\ref{tab:appendix/shmp} and Appendix~\ref{sec:appendix/SHMP/discussion}.

The summation of the neighborhood counts (the predicted canonical counts of all canonical neighborhoods) can serve as the final subgraph count prediction. The counts also show the position of patterns. But to further improve counting accuracy, we pass the neighborhood counts to the \textit{gossip propagation} stage.


\subsection{Gossip Propagation}\label{sec:gossip_propagation}

Given the count predictions $\hat{C}_c$ output by the GNN, \name uses \textbf{gossip propagation} to improve the prediction quality, enforcing different homophily and antisymmetry inductive biases for different queries.
Gossip propagation uses another GNN to model the error of neighborhood count. It uses the predicted $\hat{C}_c$ as input, and the canonical counts $C_c$ as the supervision for corresponding nodes in the target graph.

\xhdr{Motivation} To further improve the counting accuracy, we identify two inductive biases: \emph{Homophily} and \emph{Antisymmetry}.
1) \emph{Homophily}: Adjacent nodes within graphs share similar graph structures, resulting in analogous canonical counts (Figure~\ref{fig:distribution}). This phenomenon, termed \emph{homophily} of canonical counts, stands out.
2) \emph{Antisymmetry}: Nodes with similar neighborhood structures, per Definition~\ref{def:canonical_count}, exhibit higher canonical counts for those with larger node indices. See right part of Figure~\ref{fig:pipeline} for an example. Details are in Appendix~\ref{sec:appendix/homo_and_anti}.

We observe a negative correlation between \emph{Antisymmetry} ratio and \emph{Homophily} in different queries, as depicted in Figure~\ref{fig:homo_anti} in Appendix~\ref{sec:appendix/homo_and_anti}. This observation inspires us to learn this relationship within models.


The edges' direction in message passing can control the \emph{homophily} and \emph{antisymmetry} properties of the graph.
With undirected edges, message propagation is a special low-pass filter~\cite{nt2019revisiting}, enhancing the homophily property of the node values. 
With directed edges pointing from small-index nodes to large-index nodes, message propagation accumulates value in large-index nodes, which enhances the antisymmetry property. 

\begin{figure}[tb]
    \centering
    \includegraphics[width=0.35\textwidth]{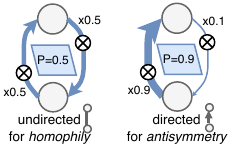}
    \vspace{-6pt}
    \caption{
        Proposed learnable gates in the gossip propagation model balance the influence of \emph{homophily} and \emph{antisymmetry} by controlling message directions.
    }
    \vspace{-8pt}
    \label{fig:gossip}
\end{figure}

\xhdr{Gossip propagation with learnable gates} To learn the edge direction that correctly emphasizes homophily or antisymmetry, we propose the gossip propagation model as shown in Figure~\ref{fig:gossip}. It multiplies a learnable gate $P$ for the message sent from the node with the smaller index, and $1-P$ for the reversed one. $P$ is learned from the query embedding. For different queries, $P$ ranges from $0$ to $1$ to balance the influence of \emph{homophily} and \emph{antisymmetry}. When $P \rightarrow 0.5$, messages from the smaller indexed node and the reversed one are weighed equally. So it simulates undirected message passing that stress \emph{homophily} by taking the average of adjacent node values. 
When the gate value moves away from 0.5, the message from one end of the edge is strengthened. For example, when $P \rightarrow 1$, the node values only accumulate from nodes with smaller indices to nodes with larger ones. So that it simulates directed message passing that stress \emph{antisymmetry} of the transitive partial order of node indices.



        
    
The messages of MPGNNs are multiplied with $g_{ji}$ on both edge directions. With learnable gates, the model can balance the effects of homophily and antisymmetry for further performance improvement. 

\begin{equation}
\begin{aligned}
    \mathbf{x}_{i}^{(k)} &= 
    \gamma^{(k)}\left(\mathbf{x}_{i}^{(k-1)}, \textsc{Agg}_{j \in N(i)} g_{ji} \cdot \phi^{(k)}\left(\mathbf{x}_{i}^{(k-1)}, \mathbf{x}_{j}^{(k-1)}, \mathbf{e}_{j, i}\right)\right)
    \\
    g_{ji} &= 
    \begin{cases}
    P & v_j\leq v_i \\
    1 - P& v_j>v_i
    \end{cases}
    \label{eq:GMP}
\end{aligned}
\end{equation}

\xhdr{Final count prediction}
The neighborhood count with gossip propagation is a more accurate estimation of the canonical count. The summation of the neighborhood counts is the unbiased estimation of subgraph count on the whole target graph as Theorem~\ref{thm:count_partition_count} states. 

\section{Experiments}\label{sec:experiments}

\begin{table}[ht]
    \centering
    \begin{tabular}{lccc}
        \toprule
        \textbf{Dataset} & \#graphs & Avg. \#nodes & Avg. \#edges \\ 
        \midrule
        \textsc{Synthetic} & 1827 & 134.91 & 381.58 \\ 
        \midrule
        \textsc{MUTAG} & 188 & 17.93 & 19.79 \\
        \textsc{COX2} & 467 & 41.22 & 43.45 \\ 
        \textsc{ENZYMES} & 600 & 32.63 & 62.14 \\ 
        \midrule
        \textsc{IMDB-BINARY} &  1000 & 19.77 & 96.53 \\
        \textsc{MSRC-21} & 563 & 77.52 & 198.32 \\
        \midrule
        \textsc{FIRSTMM-DB} & 41 & 1.3K & 3.0K \\
        \textsc{CiteSeer} & 1 & 3.3K & 4.5K \\ 
        \textsc{Cora} & 1 & 2.7K & 5.4K \\
        \bottomrule
    \end{tabular}
    \caption{Graph statistics of datasets used in experiments.}
    \vspace{-15pt}
    \label{tab:real_dataset_stats}
\end{table}

We compare the performance of \name with state-of-the-art neural subgraph counting methods, as well as the approximate heuristic method. Our evaluation showcases the scalability and generalization capabilities of \name across diverse and larger target datasets, contrasting with prior neural methods that mostly focused on smaller datasets. We also demonstrate the runtime advantage of \name compared to recent exact and approximate heuristic methods. Extensive ablation studies further show the benefit of each component of \name.

\subsection{Experimental Setup}\label{sec:exp/setup}

\xhdr{Datasets} Compared with previous neural methods, our evaluation extends to larger datasets across various domains, such as chemistry (MUTAG~\cite{debnath1991mutag}, COX2~\cite{ryan2015tudataset}), biology (ENZYMES~\cite{borgwardt2005protein}), social networks (IMDB-BINARY~\cite{yanardag2015deep}), computer vision (MSRC-21, FIRSTMM-DB~\cite{neumann2016propagation}), and citation networks (CiteSeer, Cora~\cite{mccallum2000automating}). A synthetic dataset, representing mixed graph characteristics, is also included (Table ~\ref{tab:real_dataset_stats}). Additional dataset details are in Appendix~\ref{sec:appendix/experiment_setup}.

\xhdr{Generalization framework} Our framework, trained on the Synthetic dataset with standard queries (size $3-5$), enables subgraph counting across diverse datasets and graph.

\xhdr{Baselines} \name is compared with SOTA subgraph counting GNNs: LRP~\cite{Chen2020CanGN}, DIAMNet~\cite{Liu2020NeuralSI}, DMPNN~\cite{liu2022dualMessage}, the heuristic MOTIVO~\cite{bressan2019motivo}, and exact methods VF2~\cite{cordella2004sub} and IMSM~\cite{sun2020memory}. Optimal configurations for each method are detailed in Appendix~\ref{sec:appendix/hyper_parameter} and ~\ref{sec:appendix/runtime}.

\xhdr{Evaluation metric} Evaluation utilizes mean square error (MSE) and mean absolute error (MAE) for subgraph count predictions, with MSE normalized by ground truth variance~\cite{Chen2020CanGN}.


\subsection{Neural Counting}
\label{sec:experiments/neural_counting}

\begin{table*}[tb]
    \vspace{-5pt}
    \centering
    \resizebox{\linewidth}{!}{
    \setlength\tabcolsep{3pt}
    \begin{tabular}{lccc|ccc|ccc|ccc|ccc}
        \toprule
        Dataset & \multicolumn{3}{c|}{MUTAG} & \multicolumn{3}{c|}{COX2} & \multicolumn{3}{c|}{ENZYMES} & \multicolumn{3}{c|}{IMDB-BINARY} & \multicolumn{3}{c}{MSRC-21}\\
        Query-Size & 3 & 4 & 5 & 3 & 4 & 5 & 3 & 4 & 5 & 3 & 4 & 5 & 3 & 4 & 5 \\ 
        \midrule
        \multicolumn{16}{c}{normalized MSE} \\
        \midrule
        MOTIVO & 2.9E-1 & 6.7E-1 & 1.2E+0 & 1.6E-1 & 3.4E-1 & 5.9E-1 & 1.6E-1 & 1.9E-1 & 3.0E-1 & 2.7E-2 & \textbf{3.9E-2} & \textbf{5.0E-2} & 4.8E-2 & 7.2E-2 & 9.5E-2 \\ 
        LRP & 1.5E-1 & 2.7E-1 & 3.5E-1 & 1.4E-1 & 2.9E-2 & 1.1E-1 & 8.5E-1 & 5.4E-1 & 6.2E-1 & inf & inf & inf & 2.4E+0 & 1.4E+0 & 1.1E+0 \\ 
        DIAMNet & 4.1E-1 & 5.6E-1 & 4.7E-1 & 1.1E+0 & 7.8E-1 & 7.2E-1 & 1.4E+0 & 1.1E+0 & 1.0E+0 & 1.1E+0 & 1.0E+0 & 1.0E+0 & 2.7E+0 & 1.6E+0 & 1.3E+0\\ 
        DMPNN & 6.1E+2 & 6.6E+2 & 3.0E+2 & 2.6E+3 & 2.4E+3 & 3.0E+3 & 2.9E+3 & 1.4E+3 & 1.2E+3 & 2.1E+4 & 1.3E+2 & 1.4E+2 & 1.1E+4 & 1.3E+3 & 4.1E+2 \\
        \midrule
        \name & \textbf{2.2E-3} & \textbf{7.5E-4} & \textbf{6.0E-3} & \textbf{6.6E-4} & \textbf{6.3E-4} & \textbf{4.9E-3} & \textbf{5.4E-3} & \textbf{5.9E-2} & \textbf{5.3E-2} & \textbf{8.5E-3} & 2.1E-1 & 4.5E-1 & \textbf{2.5E-3} & \textbf{3.8E-3} & \textbf{8.7E-2}\\
        \midrule
        \multicolumn{16}{c}{MAE} \\
        \midrule
        MOTIVO & 4.9E+0 & 5.1E+0 & 3.3E+0 & 8.3E+0 & 9.4E+0 & 7.3E+0 & 1.7E+1 & 2.3E+1 & 2.6E+1 & 4.7E+1 & \textbf{1.6E+2} & \textbf{6.1E+2} & 4.1E+1 & 9.5E+1 & 1.7E+2 \\ 
        LRP & 3.8E+0 & 5.1E+0 & 4.5E+0 & 9.5E+0 & 4.0E+0 & 6.3E+0 & 4.3E+1 & 4.0E+1 & 3.7E+1 & inf & inf & inf & 3.2E+2 & 4.6E+2 & 5.9E+2 \\ 
        DIAMNet & 8.3E+0 & 7.9E+0 & 4.2E+0 & 3.0E+1 & 1.7E+1 & 1.2E+1 & 5.4E+1 & 5.1E+1 & 4.0E+1 & 2.9E+2 & 8.3E+2 & 2.6E+3 & 3.4E+2 & 4.9E+2 & 6.3E+2 \\ 
        DMPNN & 6.8E+2 & 6.9E+2 & 2.4E+2 & 3.6E+3 & 4.3E+3 & 3.8E+3 & 4.8E+3 & 5.8E+3 & 6.0E+3 & 1.7E+5 & 2.2E+5 & 2.8E+5 & 3.4E+4 & 4.6E+4 & 5.7E+4 \\
        \midrule
        \name & \textbf{5.0E-1} & \textbf{1.8E-1} & \textbf{2.9E-1} & \textbf{6.1E-1} & \textbf{4.4E-1} & \textbf{7.7E-1} & \textbf{3.6E+0} & \textbf{1.1E+1} & \textbf{9.9E+0} & \textbf{2.4E+1} & 3.0E+2 & 1.6E+3 & \textbf{1.0E+1} & \textbf{2.5E+1} & \textbf{1.3E+2} \\
        \bottomrule
    \end{tabular}
    }
    \caption{Normalized MSE and MAE performance of approximate heuristic and neural methods on subgraph counting of twenty-nine standard queries.}
    \label{tab:main_tab}
    \vspace{-8pt}
\end{table*}

\xhdr{Subgraph counting} Table~\ref{tab:main_tab} highlights \name's performance in subgraph count prediction across twenty-nine standard queries of size $3-5$. It outperforms the best neural baseline and approximate heuristic method in normalized MSE by $49.7\times$ and $17.5\times$, and in MAE by $8.4\times$ and $4.1\times$ respectively. The model shows robust performance even on dense graphs which is challenging for neural method, like IMDB-BINARY. Unlike the heuristic method with exponential complexity, \name maintains linear runtime efficiency. Additional q-error metric analysis is in Appendix~\ref{sec:appendix/q_error}. 

\xhdr{Position distribution} \name innovates in pattern position prediction, achieving $3.8 \times 10^{-3}$ normalized MSE, further detailed in Appendix~\ref{sec:appendix/distribution}.

\begin{figure*}[tb]
    \centering
    \includegraphics[width=0.98\textwidth]{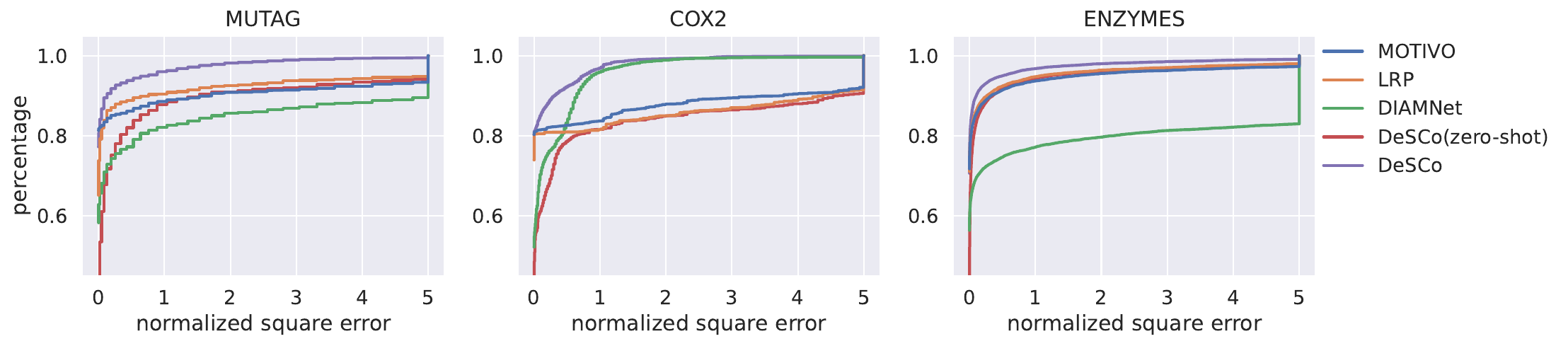}
    \vspace{-8pt}
    \caption{
    The accumulative distributions of normalized square error of large queries (size up to 13) on three target datasets. The x-axis is clipped at 5. 
    Given any square error tolerance bound (x-axis), \name has the highest percentage of predictions that meet the bound (y-axis).
    DeSCo(zero-shot) generalizes to unseen queries with competitive performance over specifically trained baselines.
    }
    \label{fig:exp/large_query}
    \vspace{-5pt}
\end{figure*}

\subsection{Scalability}\label{sec:exp/scale}

\xhdr{Large queries}
We analyze 16 frequently appearing queries for sizes $6$ to $13$ from ENZYMES (details in Appendix~\ref{sec:appendix/query_graphs}). All models, except \name(zero-shot), are fine-tuned on larger queries using the synthetic dataset. \name(zero-shot) demonstrates its capability to generalize to unseen queries. The square error distribution for each query-target pair is in Figure~\ref{fig:exp/large_query}, with numeric results in Appendix~\ref{sec:appendix/mse}.

\xhdr{Large target} In testing on large target graphs (Table~\ref{tab:larget}), \name surpasses other neural methods, handling up to $3.8\times 10^6$ and $3.3 \times 10^7$ ground truth counts on CiteSeer and Cora, respectively. LRP's results, being infinite, are excluded from the table.

\begin{table}[tb]
    \centering
    \resizebox{\linewidth}{!}{
    \setlength\tabcolsep{4pt} 
    \begin{tabular}{lccc|ccc}
    \toprule
        Dataset & \multicolumn{3}{c|}{CiteSeer} & \multicolumn{3}{c}{Cora} \\ 
        Query-Size & 3 & 4 & 5 & 3 & 4 & 5 \\ \midrule
        \multicolumn{7}{c}{normalized MSE} \\
        \midrule
        DIAMNet & 2.0E+0 & 1.5E+0 & 1.2E+0 & 1.0E+10 & 3.2E+7 & 3.7E+4 \\
        DMPNN & 9.5E+4 & 2.5E+2 & 6.8E+1 & 1.8E+5 & 1.1E+2 & 6.7E+1 \\
        \midrule
        \name & \textbf{3.5E-5} & \textbf{9.7E-2} & \textbf{1.6E-1} & \textbf{4.2E-3} & \textbf{2.1E-1} & \textbf{6.3E-2} \\
        \midrule
        \multicolumn{7}{c}{MAE} \\
        \midrule
        DIAMNet & 1.1E+4 & 6.0E+4 & 3.6E+05 & 2.1E+9 & 1.6E+9 & 8.3E+8 \\
        DMPNN & 6.1E+6 & 7.6E+6 & 8.7E+6 & 1.8E+7 & 2.4E+7 & 3.0E+7 \\
        \midrule
        \name & \textbf{6.0E+1} & \textbf{1.2E+4} & \textbf{1.1E+5} & \textbf{1.3E+3} & \textbf{7.3E+4} & \textbf{5.4E+5} \\
    \bottomrule
    \end{tabular}
    }
    \caption{Normalized MSE and MAE performance of neural methods on large targets with standard queries.}
    \label{tab:larget}
    \vspace{-8pt}
\end{table}

\subsection{Generalization Ability}
\label{sec:generalization}

\begin{figure*}[tb]
    \centering
    \includegraphics[width=0.82\textwidth]{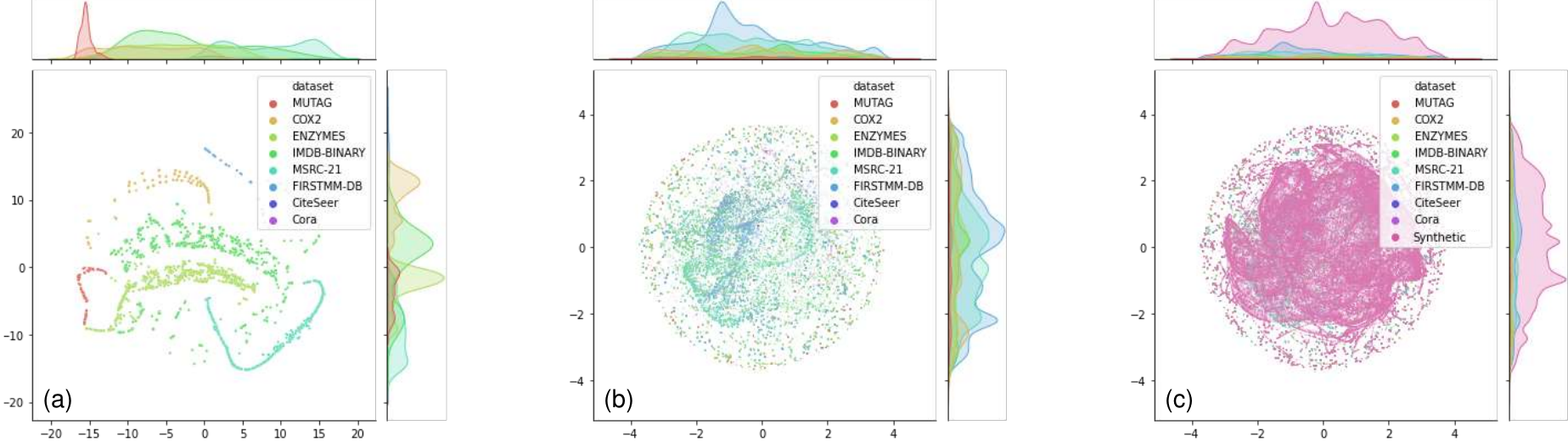}
    \vspace{-8pt}
    \caption{Visualization of statistics of diverse graph datasets. The embedding is obtained by projecting the vectors of graph statistics via t-SNE. (a) Each point represents a graph. (b) Each point represents a canonical neighborhood. (c) Canonical neighborhoods of the synthetic dataset cover most canonical neighborhoods of real-world graphs in terms of data distribution.}
    \label{fig:tsne}
    \vspace{-5pt}
\end{figure*}

\xhdr{Synthetic Dataset} Using the Synthetic dataset, we showcase \name's generalization. Real-world graphs' diversity in structure (Figure~\ref{fig:tsne} (a)) contrasts with their local substructure similarities (Figure~\ref{fig:tsne} (b)). The synthetic dataset's coverage of real-world graph characteristics (Figure~\ref{fig:tsne} (c)) confirms \name's training effectiveness and generalizability.

\xhdr{Generalization} \name, pre-trained on the Synthetic dataset and tested on varied real-world datasets, demonstrates superior accuracy and generalization compared to models trained on existing datasets (Table~\ref{tab:general_small}). This underscores its robustness across different domains.

\begin{table}[tb]
    \hspace*{-8pt}
    \footnotesize
    \setlength\tabcolsep{2pt}

    \begin{tabular}{lccc|ccc|ccc}
        \toprule
        Test-Set & \multicolumn{3}{c|}{MUTAG} & \multicolumn{3}{c|}{MSRC-21} & \multicolumn{3}{c}{FIRSTMM-DB} \\ 
        Query-Size & 3 & 4 & 5 & 3 & 4 & 5 & 3 & 4 & 5 \\ 
        \midrule
        Existing  & 6.5E-3 & 3.4E-3 & 8.7E-2 & 1.1E+1 & 1.9E+0 & 1.1E+0 & 1.1E-1 & 1.1E-1 & 1.6E-1  \\
        Synthetic & \textbf{2.3E-3} & \textbf{8.4E-4} & \textbf{6.5E-3} & \textbf{2.5E-3} & \textbf{3.8E-3} & \textbf{8.7E-2} & \textbf{2.1E-3} & \textbf{3.6E-2} & \textbf{5.4E-2}  \\
        \bottomrule
    \end{tabular}%
    \caption{Normalized MSE performance with different training datasets. When pre-training on existing datasets, MSRC-21 uses MUTAG; CiteSeer uses Cora; FIRSTMM-DB uses CiteSeer.}
    \label{tab:general_small}
    \vspace{-12pt}
\end{table}

\subsection{Ablation Study}

\begin{figure}[tb]
    \vspace{-3pt}
    \centering
    \includegraphics[width=0.45\textwidth]{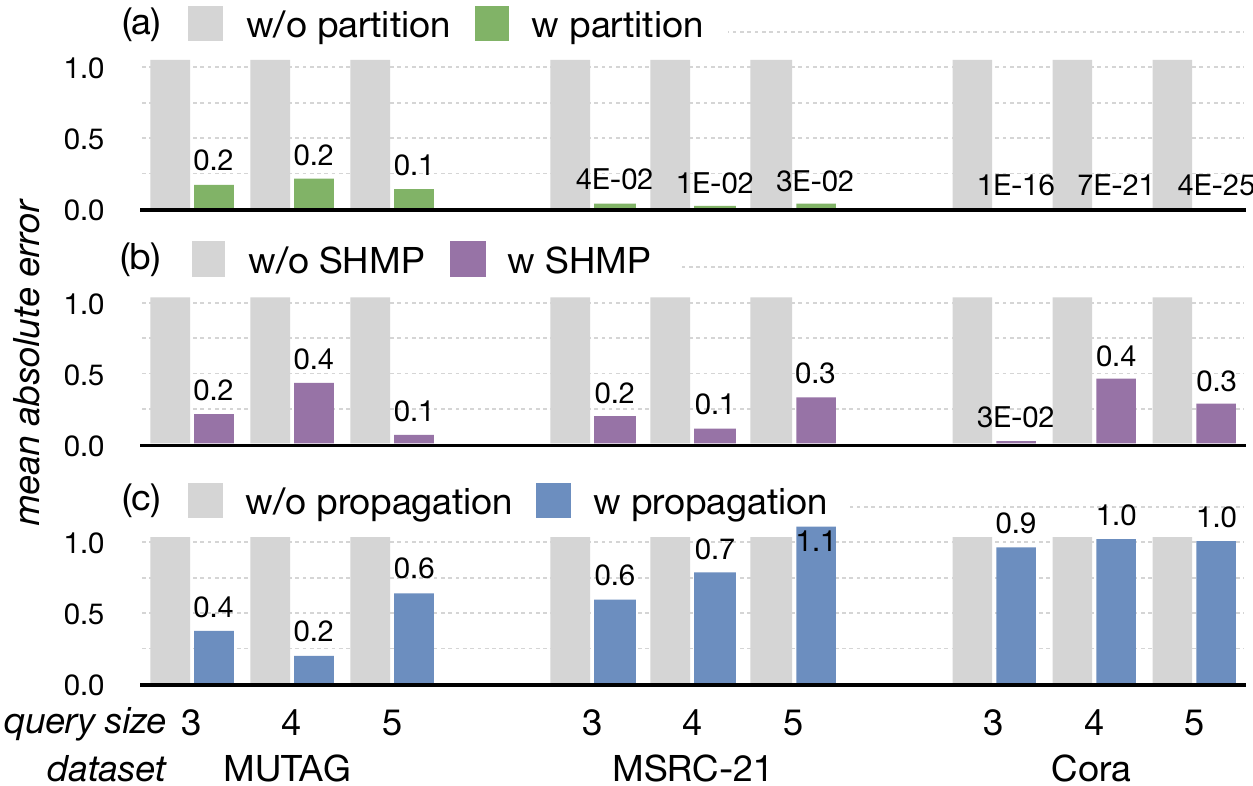}
    \vspace{-5pt}
    \caption{MAE performance with and without canonical partition, SHMP and gossip propagation.}
    \label{fig:exp/ablation}
    \vspace{-10pt}
\end{figure}

In assessing \name's components, the ablation study reveals significant contributions of each part. 
We demonstrate the MAE results on three datasets (Figure~\ref{fig:exp/ablation}) and the geometric mean of normalized MSE on eight datasets (Figure~\ref{fig:intro/benefit}), supported by numeric data in Appendix ~\ref{sec:appendix/numeric_result}.

\xhdr{Ablation of canonical partition} Removing the canonical partition and training \name for subgraph count on whole targets (like other neural baselines) indicates the partition's vital role in error reduction and \name's superiority over existing neural methods (Figure~\ref{fig:intro/benefit}). Canonical partition in \name brings a $1.3 * 10^{10} \times$ improvement in normalized MSE and $8.8 * 10^4 \times$ in MAE.

\xhdr{Ablation of SHMP} SHMP enhances GraphSAGE's performance by transitioning to heterogeneous message passing, using triangles as the categorizing subgraph (Figure~\ref{fig:SHMP}). SHMP reduces the normalized MSE by $27\times$ and MAE by $5.8\times$ over GraphSAGE. Further more, when compared with expressive GNNs, including GIN and ID-GNN, SHMP demonstrate a $24 \times$ and $14\times$ reduction in normalized MSE, as well as a $5.3 \times$, $3.9 \times$ reduction MAE, as detailed in Table~\ref{tab:appendix/shmp}.

\xhdr{Ablation of gossip propagation} Comparing direct summation of neighborhood counts with summation post-gossip propagation highlights its effectiveness. Gossip propagation further reduces normalized MSE and MAE by $1.8\times$ and $1.4\times$, respectively.

\subsection{Runtime Comparison}
\begin{figure}[tb]%
    \centering
    \includegraphics[width=0.47\textwidth]{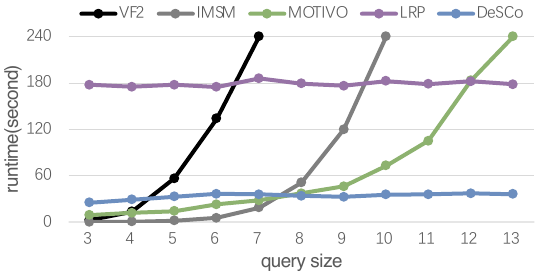}
    \vspace{-7pt}
    \caption{The runtime comparison between exact, heuristic approximate, neural methods and \name. All tested on the ENZYMES dataset.}
    \label{fig:exp/runtime}%
    \vspace{-4pt}
\end{figure}

Figure~\ref{fig:exp/runtime} illustrates the runtime of each method under a four-minute limit. Exact methods VF2 and IMSM exhibit exponential runtime increases due to the \#P hard nature of subgraph counting. For the approximate heuristic method MOTIVO, exponential growth mainly stems from its coloring phase. In contrast, neural methods LRP and \name show polynomial scalability. \name achieves a 5.3$\times$ speedup over LRP, as it avoids heavy node feature permutations. Further runtime analysis is available in Appendix~\ref{sec:appendix/runtime}.

\section{Conclusion}\label{sec:conclusion}

We propose \name, a neural network based pipeline for generalizable and scalable subgraph counting. With canonical partition, subgraph-based heterogeneous message passing, and gossip propagation, \name accurately and efficiently predicts counts for both large queries and targets. It demonstrates magnitudes of improvements in mean square error. It additionally provides the important position distribution of patterns that previous works cannot. 

\begin{acks}
This work was supported by National Natural Science Foundation of China (No. U19B2019, 62325405, U21B2031, 61832007, 62104128, 62204164), and Beijing National Research Center for Information Science and Technology (BNRist), and Tsinghua-Meituan Joint Institute for Digital Life.
\end{acks}

\clearpage
\bibliography{ref}
\bibliographystyle{ACM-Reference-Format}

\section*{Ethical Considerations}

In the realm of graph analysis, \name stands as a fundamental tool rather than a specific application-driven solution. While the direct potential for \name to induce negative societal impacts is minimal, it remains prudent to acknowledge and address potential adverse outcomes.

\xhdr{Accuracy}
Similar to other non-exact counting methods, \name cannot ensure absolute prediction correctness. Despite thorough testing on extensive real-world datasets, which has showcased significant error reductions and exceptional generalization capabilities, the potential for inaccurate predictions, especially for outlier graphs, remains a possibility. Therefore, it's advisable to exercise caution and validate basic graph statistics, such as maximum degree, before applying the \name method.

\xhdr{Privacy}
\name introduces a breakthrough in accurately counting large subgraphs, previously unattainable. Moreover, it reveals the positional distribution of these counts. As subgraph counting finds applications in recommendation systems, social network analysis, and other domains, there's potential for corporations and governments to glean intelligence that was once inaccessible. This advancement could inadvertently compromise user privacy if not subjected to proper oversight. 
To mitigate this, it's essential to consider enforcing relevant regulations should corresponding technologies be developed.

\clearpage
\appendix
\section{Canonical Partition} 
\label{sec:appendix/prof_canonical}

\subsection{Proof of Lemma~\ref{lm:count_canonical_count}}
\label{sec:appendix/prof_lemma4d1}

\begin{proof}
Following the notions from Section~\ref{sec:preliminary}, given a query graph $G_q$ and a target graph $G_t$, the node-induced count is defined as the number of $G_t$'s node-induced subgraph, pattern, $G_p$ that is isomorphic to $G_q$. We denote the set of all $G_p$ as $\mathbb{M}$.

\begin{gather}
    \mathbb{M} = \{ G_p \subseteq G_t|G_p \cong G_q \} \label{eq:def_m} \\
    \mathcal{C}(G_q,G_t) = |\mathbb{M}| \label{eq:graph_count_init}
\end{gather}

Assuming that $G_q$ has k nodes. Then, under the node-induced definition, given $G_t$, we can use the k-node set $V_p=\{v|v \in G_p\}$ of $G_p$ to represent the pattern.

We can decompose the set of all patterns $\mathbb{M}$ into many subsets $\mathbb{M}_c$ based on the maximum node index of each $G_p \in \mathbb{M}$.

\begin{equation}
    \mathbb{M}_c = \{ G_p \subseteq G_t|G_p \cong G_q, \max_{I}V_p = c \}
    \label{eq:M_c}
\end{equation}

This maximum-index decomposition is exclusive and complete: every $G_p$ has a single corresponding maximum node index. So we have the following properties:

\begin{gather}
    \forall c \neq j, \mathbb{M}_c \cap \mathbb{M}_j = \varnothing \\
    \mathbb{M} = \bigcup_{c=0}^{|V|-1} \mathbb{M}_c
\end{gather}

Thus, the node-induced count in Equation~\ref{eq:graph_count_init} can be rewritten using the inclusion-exclusion principle.

\begin{equation}
\begin{aligned}
    \mathcal{C}(G_q,G_t) 
    &= \left| \bigcup_{c=0}^{|V|-1} \mathbb{M}_c \right|\\
    &= \sum_{c=0}^{|V|-1} |\mathbb{M}_c|
    + \sum_{k=1}^{|V|-1}(-1)^k \left( \sum_{0\leq i_0\leq \cdots i_k < |V|}\left|\mathbb{M}_{i_{0}} \cap \cdots \cap \mathbb{M}_{i_{k}} \right| \right)\\
    &= \sum_{c=0}^{|V|-1} |\mathbb{M}_c|
\end{aligned}
\label{eq:prof_count_canonical_count}
\end{equation}

According to the definition of canonical count in  Equation~\ref{eq:canonical_count}, $\mathcal{C}_c(G_q,G_t,v_c)=|\mathbb{M}_c|$. Thus, Lemma~\ref{lm:count_canonical_count} is proven with Equation~\ref{eq:prof_count_canonical_count}.

\end{proof}

\subsection{Proof of Theorem~\ref{thm:count_partition_count}}
\label{sec:appendix/theo1}

\begin{proof}

By the definition of $\mathbb{M}_c$ in Equation~\ref{eq:M_c}, we have a corollary.

\begin{corollary}
Denote $v_c$'s index as $c$, $\mathcal{D}$ as the shortest path length between two nodes. Any graph in $\mathbb{M}_c$ has node $v_c$ and has the same graph-level property with $G_q$, e.g., diameter. 
\begin{equation}
    \forall G_p \in \mathbb{M}_c, v_c \in V_p, \max_{v_i, v_j \in V_p}\mathcal{D}(G_p, v_i, v_j) = \max_{v_i, v_j \in V_q}\mathcal{D}(G_q, v_i, v_j)
\end{equation}
\label{cor:share v_c}
\end{corollary}

The distance between $v_c$ and any nodes of $G_p$ in $\mathbb{M}_c$ is bounded by $\max_{v_i, v_j \in V_q}\mathcal{D}(G_q, v_i, v_j)$ as shown in corollary \ref{cor:share v_c}. So we can further know that graphs in $\mathbb{M}_c$ are node-induced subgraphs of $v_c$'s d-hop ego-graph.

\begin{equation}
\begin{aligned}
    & \forall G_p \in \mathbb{M}_c, \exists G_{d-ego} \subseteq G_t, V_{d-ego} = \{ v_i \in V_t|\mathcal{D}(G_t,v_i,v_c) \leq d\} \\
    & \operatorname{ s.t. } G_p \cong G_{d-ego}
    \label{eq:d-ego}
\end{aligned}
\end{equation}

Given Equation~\ref{eq:M_c}, it is also clear that all graphs in $\mathbb{M}_c$ have smaller node indices than $c$.

\begin{equation}
    \forall G_p \in \mathbb{M}_c, \exists G_{small} \subseteq G_t, V_{small} = \{ v_i \in V_t|I_i \leq I_c\} \operatorname{ s.t. } G_p \cong G_{small}
    \label{eq:smaller_than_c}
\end{equation}

With Equation~\ref{eq:d-ego} and \ref{eq:smaller_than_c}, we know that all the graphs in $\mathbb{M}_c$ are subgraphs of $\mathcal{P}(G_t, v_c, d)$ defined in Equation~\ref{eq:canonical_neighborhood}. Thus, with respect to Equation~\ref{eq:def_m}, we can redefine $\mathbb{M}_c$ as follows.

\begin{equation}
    \mathbb{M}_c = \{ G_p \subseteq \mathcal{P}(G_t, v_c, d)|G_p \cong G_q, \max_{V_p}I = c \}
    \label{eq:M_c_P}
\end{equation}

Combining Equation~\ref{eq:prof_count_canonical_count} with Equation~\ref{eq:M_c_P}, Theorem~\ref{thm:count_partition_count} is proven.
\end{proof}

\subsection{Implementation of Canonical Partition}
\label{sec:appendix/implement_canonical_partition}

\begin{algorithm}
\caption{Index-restricted breadth-first search}
\label{alg:id-bfs}
\begin{algorithmic}
    \State $V_c \gets \{v_c\}$, $V_{front} \gets \{v_c\}$, $depth \gets 0$
    \While{$depth < d$}
        \State $ V_{add} \gets \{v | v \in \bigcup_{v_i \in V_{front}}  \{v_j|(v_i,v_j) \in E_t\}, v \leq v_c \} $
        \State $ V_{front} \gets V_{add} \setminus V_c $
        \State $V_c \gets V_c \cup V_{front}$
        \State $depth \gets depth + 1$  
    \EndWhile
    \State $G_c \gets (V_c, E_c) \operatorname{ s.t. } G_c \subseteq G_t $
\end{algorithmic}
\label{alg:partition}
\end{algorithm}

The canonical partition is implemented using an index-restricted breadth-first search (BFS). Compared with regular BFS, it restricts the frontier nodes to have smaller indices than that of the canonical node. The time complexity of canonical partition equals the BFS on each neighborhood $G_n=(V_n, E_n)$, which is $\sum O(V_n+E_n) = O(V_t\times(\bar{V}_n+\bar{E}_n))$.

\subsection{Complexity Benefit of Canonical Partition}
\label{sec:appendix/complexity_benefit}

\begin{figure}[ht]
    \centering
    \includegraphics[width=0.5\textwidth]{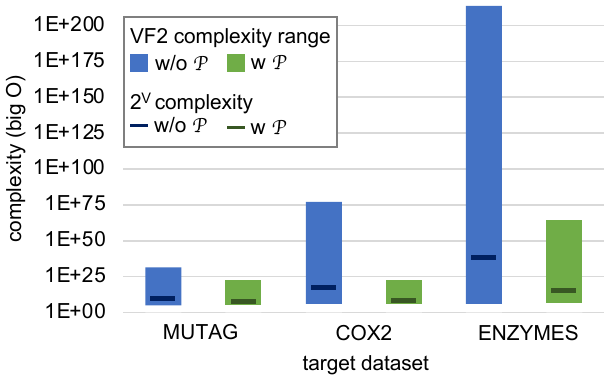}
    \caption{The complexity of subgraph counting with and without canonical partition on different target datasets. The complexity for the VF2 exact subgraph counting method is $O(V^2)$ to $O(V!\times V)$. The $O(2^V)$ complexity estimates the empirically observed average complexity.}
    \label{fig:appendix/complexity}
\end{figure}
We discuss the computational benefit of the canonical partition method in this section.

\xhdr{Search space reduction} 
Canonical partition uses the divide-and-conquer scheme to bring about drastic search space reduction.
We denote the complexity of searching and counting all subgraphs on size $V_t$ target graph as $S(V_t)$. The Canonical partition divides the original problem into subproblems with the total search space of $\sum_{i \in V_t} S(V_{n_i})$, where $V_{n_i}$ stands for the size of canonical neighborhoods. 
Thanks to the sparse nature of real-world graphs, $V_{n_i}$s are generally small, even for huge target graphs. So with canonical partition, the search space is drastically reduced.

We conduct experiments on real-world graphs to show how canonical partition fundamentally reduces the search space. Figure~\ref{fig:appendix/complexity} shows the computational complexity with different assumptions in the form of $S$. VF2~\cite{cordella2004sub} claims that the asymptotic complexity for the problem ranges from $O(V^2)$ to $O(V!\times V)$ in best and worst cases. Under such assumptions of $S$, the average worst-case complexity is reduced by a factor of $1/10^{70}$ with canonical partition, while the average best-case complexity stays in the same magnitude. 
Empirically, we observe exponential runtime growth of the subgraph counting problem. Thus, under the assumption that $S(V)=2^V$, the average complexity is also reduced drastically by a factor of $1/10^{11}$ with canonical partition.

\xhdr{Redundant match elimination}
Canonical partition, along with the canonical count definition, eliminates the redundant automorphic match of the query graph. Previous works~\cite{mawhirter2019automine,shi2020graphpi} have shown that the automorphism of the query graph can cause a large amount of redundant count. 
For example, the triangle query graph $G_q$ has three symmetric nodes. We denote the triangle pattern as $G_p \subseteq G_t$ and the bijection $\mathbb{R}^3 \mapsto \mathbb{R}^3$ as $f:(v_{p_0},v_{p_1},v_{p_2}) \mapsto (v_{q_0},v_{q_1},v_{q_2})$. For the same pattern, there exist six bijections $\{f:(v_{p_0},v_{p_1},v_{p_2}) \mapsto (v_{q_i},v_{q_j},v_{q_k}) | (i,j,k) \in \text{Perm}(1,2,3) \}$ where $\text{Perm}(x,y,z)$ denotes all $3!$ permutations of $(x,y,z)$. 

Canonical partition eliminates such redundant bijections by adding asymmetry, the canonical node. As discussed in Equation~\ref{eq:canonical_count}, by attributing the count to only one canonical node, the bijection $f_c$ can be rewritten as a $\mathbb{R}^3 \mapsto \mathbb{R}$ function, $f_c:(v_{p_0},v_{p_1},v_{p_2}) \mapsto \max_I(v_{q_0},v_{q_1},v_{q_2})$. It means that each query corresponds to only one bijection instead of six, thus preventing double counting and reducing the computational complexity.

\xhdr{Reduction for the variation of counts} Canonical partition also reduces the variation of counts, which makes the regression task easier for the neural network as discussed in Section~\ref{sec:intro}. The detailed statistics of the range of counts (maximum count minus minimum count) are shown in Appendix~\ref{sec:appendix/query_graphs}. The canonical partition reduces the range of counts to $1/3$ on average in Figure~\ref{fig:appendix/std_query_example}.

\section{Expressive Power of SHMP}

\begin{figure}[ht]
    \centering
    \includegraphics[width=0.5\textwidth]{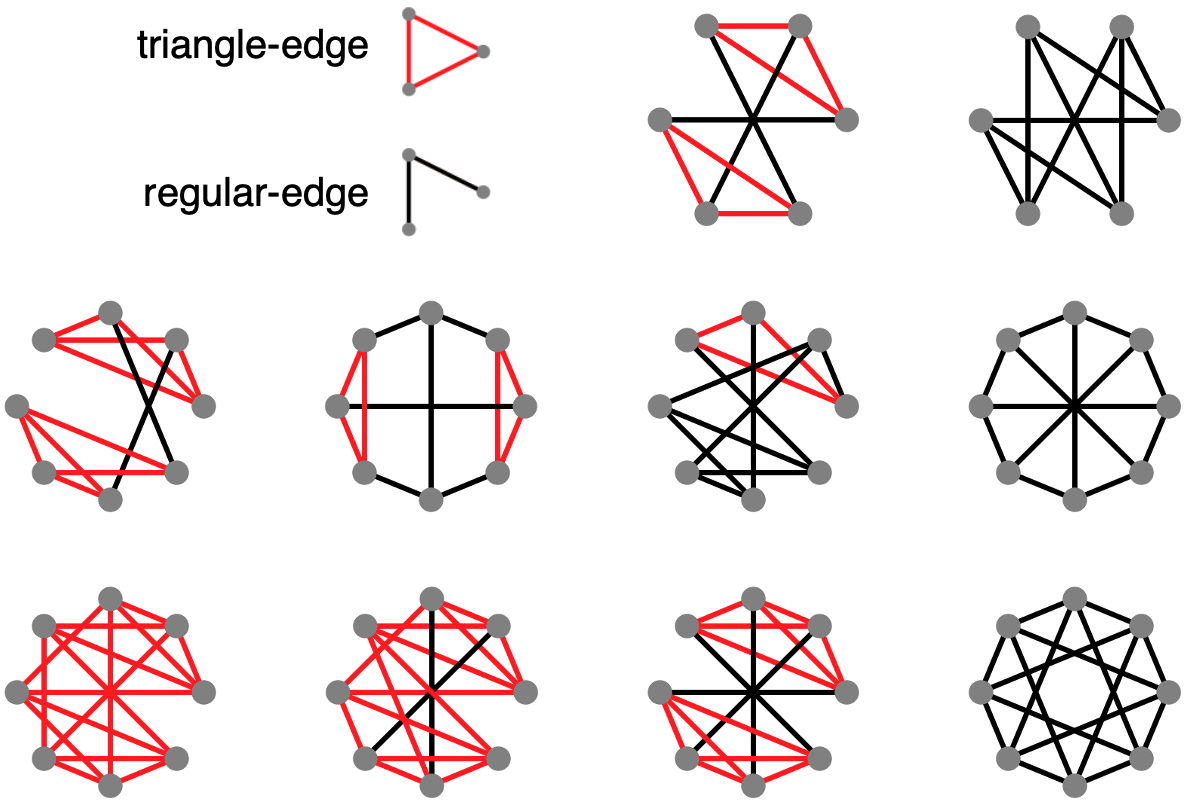}
    \caption{Examples of 1-SHMP distinguishable graphs. 
    Any graph pair of each row cannot be distinguished by the 1-WL test. 
    While with one layer of triangle-based SHMP, the histogram of the triangle-edge can distinguish all these graph pairs.}
    \label{fig:appendix/d-regular}
\end{figure}

\subsection{Theoretical Comparison with Regular Message Passing}
\label{sec:appendix/prof_expressive_power}

Previous work~\cite{xu2018powerful} has shown that the expressive power of existing message passing GNNs is upper-bounded by the 1-WL test, and such bound can be achieved with the Graph Isomorphism Network (GIN). We prove the expressive power of SHMP with the following Lemma.

\begin{lemma}
    The SHMP version of GIN has stronger expressive power than the 1-WL test.
\end{lemma}

By setting $\forall \phi_h^{k} = \phi^{k}$ and $\textsc{Agg}'=\textsc{Agg}$, SHMP from Equation~\ref{eq:SHMP} becomes an instance of GIN, which proves that SHMP-GIN is at least as expressive as GIN or the 1-WL test. The examples in Figure~\ref{fig:appendix/d-regular} and Table~\ref{tab:appendix/d-regular-rate} further prove that one layer of triangle-based SHMP-GIN can distinguish certain graphs that the 1-WL test cannot. 
Thus, SHMP-GIN has stronger expressive power than the 1-WL test, exceeding the upper bound of regular message passing neural networks.

\subsection{Experiments on Regular Graphs}
\label{sec:appendix/SHMP/regular_graph}

To further illustrate the expressive power of SHMP, we show the number of graph pairs that are WL indistinguishable but SHMP distinguishable in Table~\ref{tab:appendix/d-regular-rate}. 
We collect all the connected, d-regular graphs of sizes six to twelve from the House of Graphs~\cite{brinkmann2013house}.
Among these 157 graphs, 654 pairs of graphs are indistinguishable by the 1-WL test, even with infinite iterations. In comparison, only 208 pairs are indistinguishable by the triangle-based SHMP with a single layer. 
So 68\% of typical fail cases of the 1-WL test are easily solved with SHMP.
Some examples are shown in Figure~\ref{fig:appendix/d-regular}.

\begin{table*}[ht]
    \centering
    \begin{tabular}{lccccccc}
    \toprule
        Graph Size & 6 & 7 & 8 & 9 & 10 & 11 & 12 \\ 
        Number of Graphs & 5 & 4 & 15 & 10 & 30 & 5 & 88 \\
        Number of Graph Pairs & 10 & 6 & 105 & 45 & 435 & 10 & 3828 \\
        \midrule
        WL Indistinguishable & 1 & 1 & 19 & 13 & 64 & 1 & 555 \\
        SHMP Indistinguishable & 0 & 1 & 4 & 4 & 26 & 0 & 173 \\ 
        Error Reduction & 100.0\% & 0.0\% & 78.9\% & 69.2\% & 59.4\% & 100.0\% & 68.8\% \\ 
    \bottomrule
    \end{tabular}
    \caption{The number of indistinguishable d-regular graph pairs for the WL-test and SHMP.}
    \label{tab:appendix/d-regular-rate}
\end{table*}

\subsection{Discussion on Substructure Enhanced GNNs}
\label{sec:appendix/SHMP/discussion}

Previous substructure enhanced GNNs~\cite{Morris2019WeisfeilerAL, Nikolentzos2020khopGN} focus on the idea of high-order abstractions of the graph. 
However, the direct instantiation of all high-order substructures poses significant runtime overhead, which is unfriendly for the large-scale subgraph counting problem.
For example,~\cite{Morris2019WeisfeilerAL} has to add $\binom{n}{k}$ nodes to represent the corresponding k-order substructure of an n-node target graph. 
This results in massive memory overhead and heavy message passing computation. 
Though both of them use the three-node substructure information, experiments show that the five-layer \name is 3.51$\times$ faster than the five-layer 1-2-3-GNN~\cite{Morris2019WeisfeilerAL} when embedding the same COX2 dataset.
In contrast, DeSCo's subgraph-based heterogeneous message passing (SHMP) focuses on the idea of distinguishing different local graph structures. By categorizing the messages on the original graph, \name efficiently uses the same amount of message passing computation as traditional MPGNNs, while providing better expressive power.

\section{Homophily and Antisymmetry Analysis of Gossip Propagation} \label{sec:appendix/homo_and_anti}

\begin{figure}[t]
    \centering
    \includegraphics[width=0.4\textwidth]{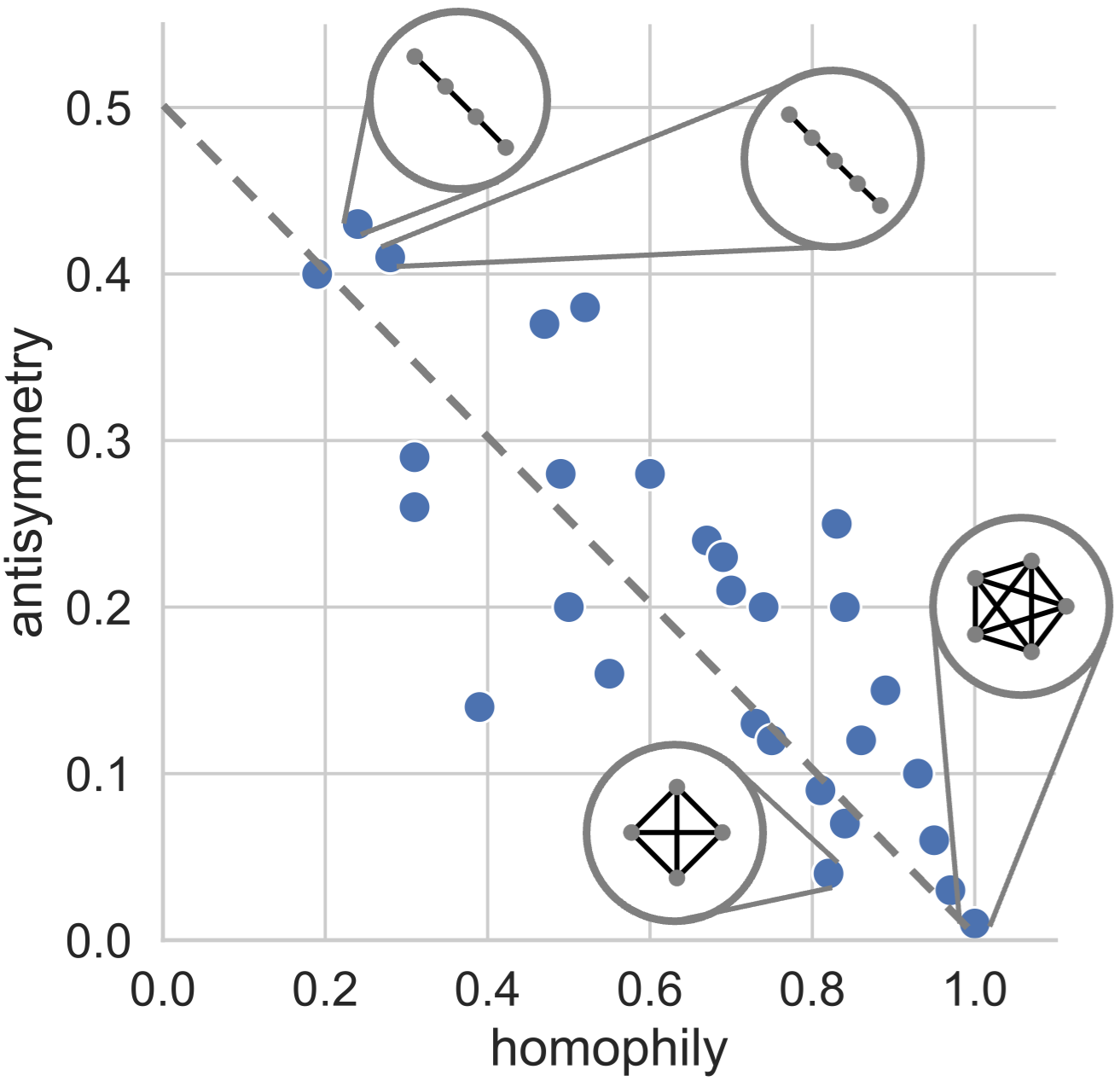}
    \vspace{-5pt}
    \caption{\emph{Homophily} and \emph{antisymmetry} of different queries, measured by the homophily ratio and the index-count correlation, respectively. The two inductive biases are negatively correlated.}
    \label{fig:homo_anti}
    \vspace{-10pt}
\end{figure}

\xhdr{Explanation}
\revision{1) \emph{Homophily}. Since the neighborhoods of adjacent nodes share much common graph structure, they tend to have similar canonical counts as shown in Figure~\ref{fig:distribution}. It is called the \emph{homophily} of canonical counts.
2) \emph{Antisymmetry}. As mentioned after Definition~\ref{def:canonical_count}, for nodes with similar neighborhood structures, the one with a larger node index has a larger canonical count, resulting in \emph{antisymmetry} of canonical counts. See the example target graph with neighborhood counts in Figure~\ref{fig:pipeline}. The nodes $(0,1,2)$ and $(3,5,6)$ are both triangles, yet $(3,5,6)$ have larger node indices, thus larger canonical counts.
Figure~\ref{fig:homo_anti} further show that \emph{homophily} and \emph{antisymmetry} have a negative correlation for different queries, which inspires us to learn such negative correlation to improve counting accuracy.}

\xhdr{Example and observation}
The \emph{homophily} and the \emph{antisymmetry} are two important inductive biases for the canonical count. The target graph in Figure~\ref{fig:pipeline} serves as a vivid example. 
The numbers in the green square indicate the canonical count value of each node. 
On the one hand, note that the adjacent nodes 3, 5, and 6 have the same count value of 2. Adjacent nodes 0, 1, and 2 also have the same value, 0. This homophily inductive bias suggests that taking an average of the adjacent node values can reduce the prediction error of individual nodes.
On the other hand, though node 1 and node 5 have similar neighborhood graph structures, node 5, with a larger node index, has a higher canonical count value. It corresponds to the definition of canonical count as discussed in Section~\ref{sec:pipeline/canonical_count}. This antisymmetry inductive bias suggests that the embedding phase for two structurally similar nodes with different node indices should also be different.

\xhdr{Quantization}
We quantify the \emph{homophily} and the \emph{antisymmetry} inductive biases. 
For homophily, we treat the canonical count as the node label and use the homophily ratio from~\cite{zhu2021heterophily} to quantify how similar the count is between adjacent nodes. The homophily ratio ranges from 0 to 1. The higher the homophily ratio is, the more similar the labels will be between adjacent nodes. 
For antisymmetry, we use the Pearson correlation coefficient $r$ ~\cite{pearson} between the node index and its canonical count as the quantification metric. 
We quantify the different \emph{homophily} and \emph{antisymmetry} for the standard queries on the ENZYMES target graph. 

\xhdr{Key insight}
As shown in Figure~\ref{fig:homo_anti}, the key insight is that \emph{homophily} and the \emph{antisymmetry} generally have negative correlation $r=-0.82$. So the emphasis on one should suppress the other. Based on such observation, we design the gossip propagation model with learnable gates to imitate the mutually exclusive relation between the two inductive biases for different queries. As shown in Figure~\ref{fig:gossip}, the proposed learnable gate balances the influence of \emph{homophily} and \emph{antisymmetry} by controlling the direction of message passing. The gate value is trained to adapt for different queries to imitate different extents of \emph{homophily} and \emph{antisymmetry}.

\section{Experimental Setup}
\label{sec:appendix/experiment_setup}

\subsection{Synthetic Dataset}
\label{sec:appendix/experiment_setup/synthetic_dataset}

The \name can be pre-trained once and be directly applied to any targets. So we generate a synthetic datasets. 
Synthetic dataset has 1827 graphs. It is designed to generally cover various graph datasets.
We use the exact counting method to generate the ground truth counts of all twenty-nine \emph{standard queries} (queries of sizes 3, 4, 5) of this synthetic graphs.

The synthetic dataset generates each graph with a generator from the generator pool. 
The pool consists of six different graph generators: the Erdős-Rényi (ER) model~\cite{erdHos1960evolution}, the Watts-Strogatz (WS) model~\cite{watts1998collective}, the Extended Barabási-Albert (Ext-BA) model~\cite{albert2000topology}, the Power Law (PL) cluster model~\cite{holme2002growing}, Barabási-Albert (BA) model~\cite{barabasi1999emergence} and the random graph generator (gnm-random-graph) from networkx~\cite{hagberg2008networkx}.

The Synthetic dataset generation process first generates a set of expected \#node, \#edge pairs as the jobs. The jobs are then randomly assigned to the six generators. Each generator tries to generate the expected graph. The generator sets its parameters so that, from the perspective of probability, the expected \#node, \#edge confirms the assigned job. 
For the \#node, \#edge pairs, the Synthetic dataset uniformly generates 1380 jobs with nodes ranging from 10 to 59 and the average degree ranging from 1 to 12. It then uniformly generates 447 jobs with nodes ranging from 60 to 800 and the average degree ranging from 1 to 3.
For the ER model, parameter $p=2m/(n(n-1))$ where $n$ and $m$ denotes the number of nodes and edges.
For the WS model, the parameter $k=2m/n$. We set $p=0.1$ for the model.
For the Ext-BA model, the parameter $p=(m-\lfloor m/n \rfloor n )/ n$. We set $q=0.1$ for the model.
For the power model, parameter $k=\lfloor(n-\sqrt{n^2-4m})/2\rfloor$, $p=(m-(n-k)k)/((k-1)(n-k))$.

\begin{figure*}
    \centering
    \includegraphics[width=\textwidth]{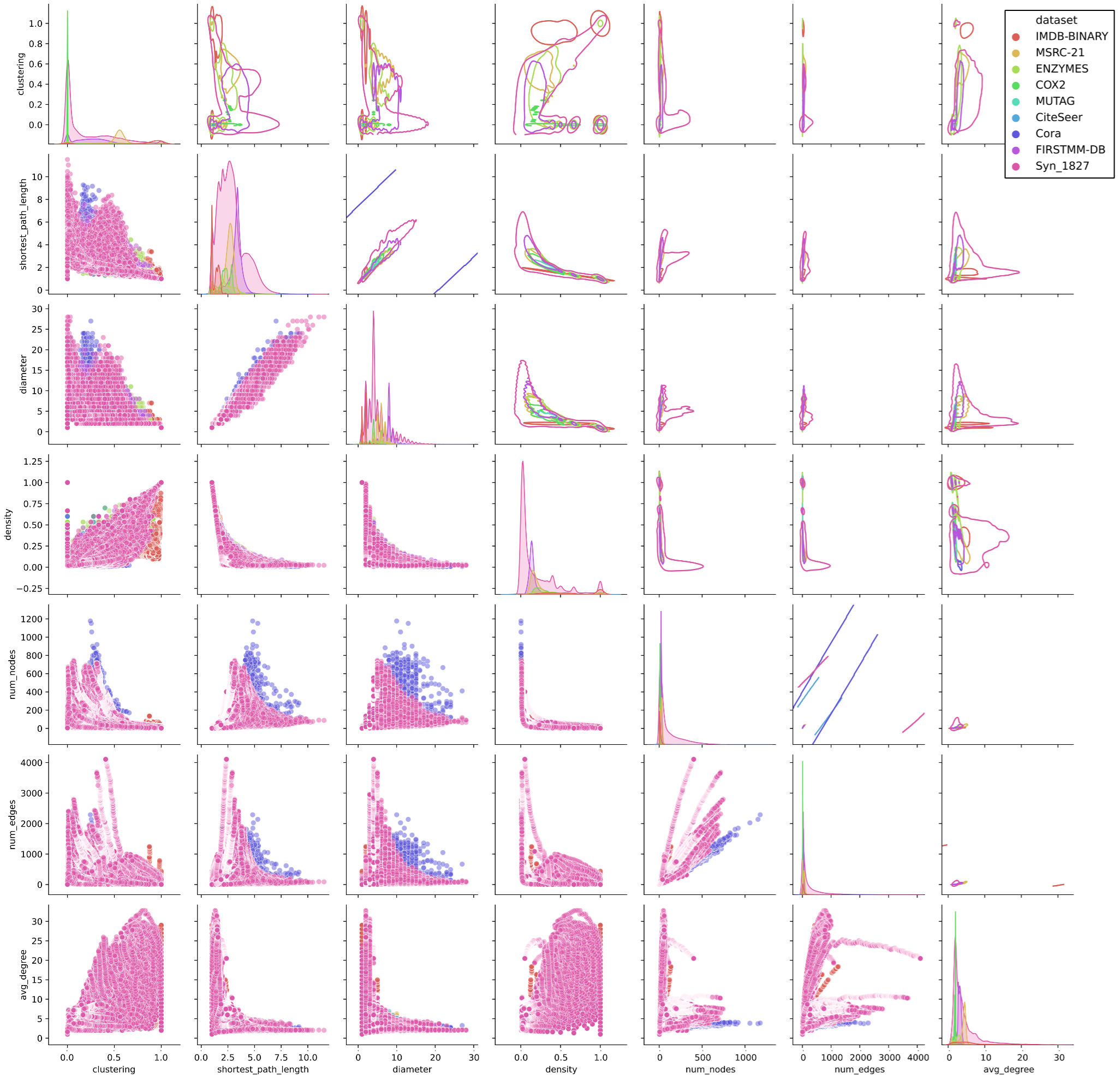}
    \caption{The canonical neighborhoods' statistics of real-world datasets and the synthetic dataset. The synthetic dataset covers most of the real-world datasets, providing a strong foundation for \name's generalization ability.}
    \label{fig:syn_neigh_comprehensive}
\end{figure*}

\subsection{Query Graphs}
\label{sec:appendix/query_graphs}

\begin{figure*}[ht]
    \centering
    \includegraphics[width=0.8\textwidth]{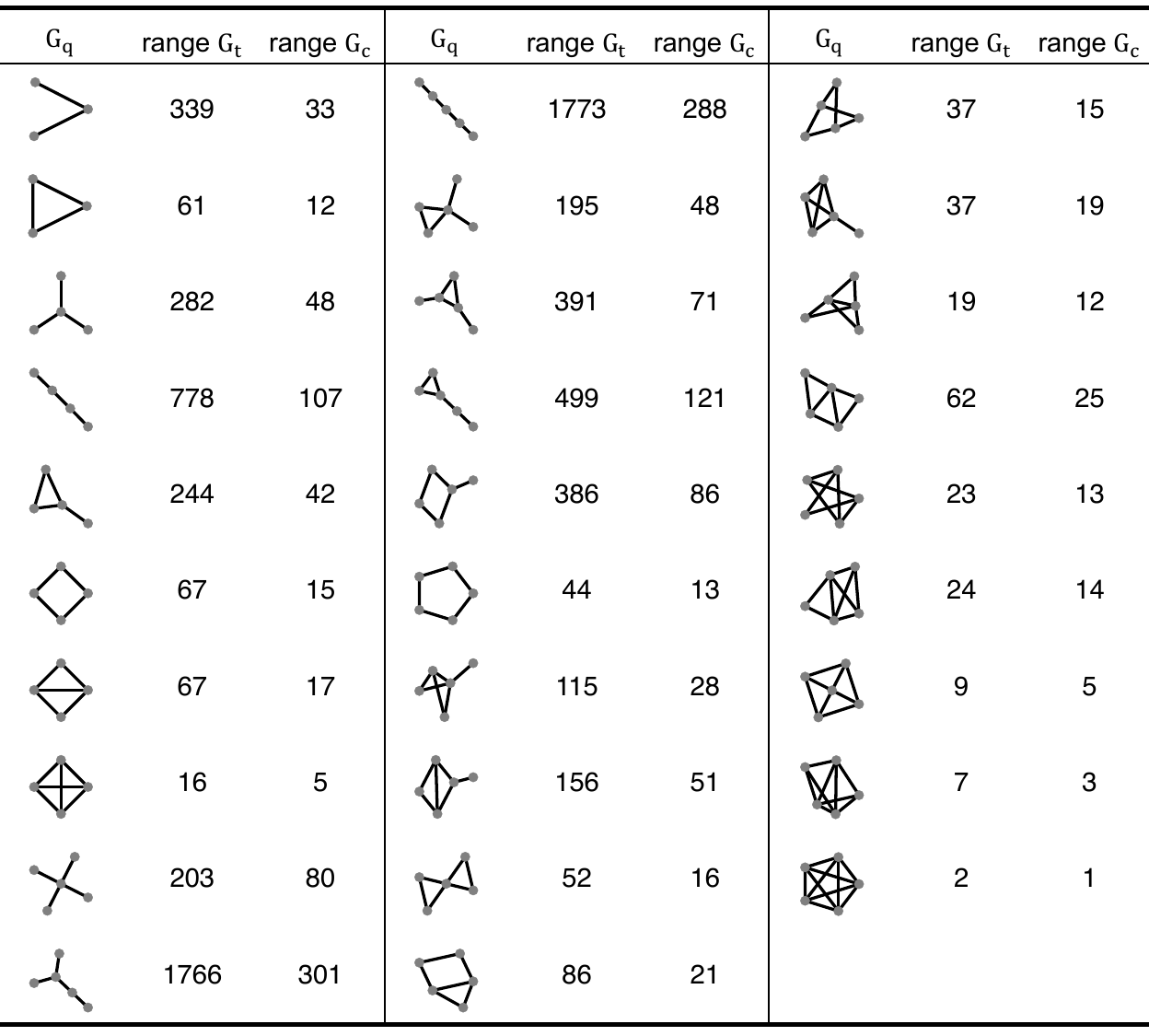}
    \vspace{-10pt}
    \caption{The \emph{standard query graphs}, the range of the subgraph counts on target graphs $G_t$, and the range of the canonical counts on neighborhoods $G_c$. The statistics come from the ENZYMES dataset.
    }
    \label{fig:appendix/std_query_example}
\end{figure*}

\begin{figure*}[ht]
    \centering
    \includegraphics[width=0.8\textwidth]{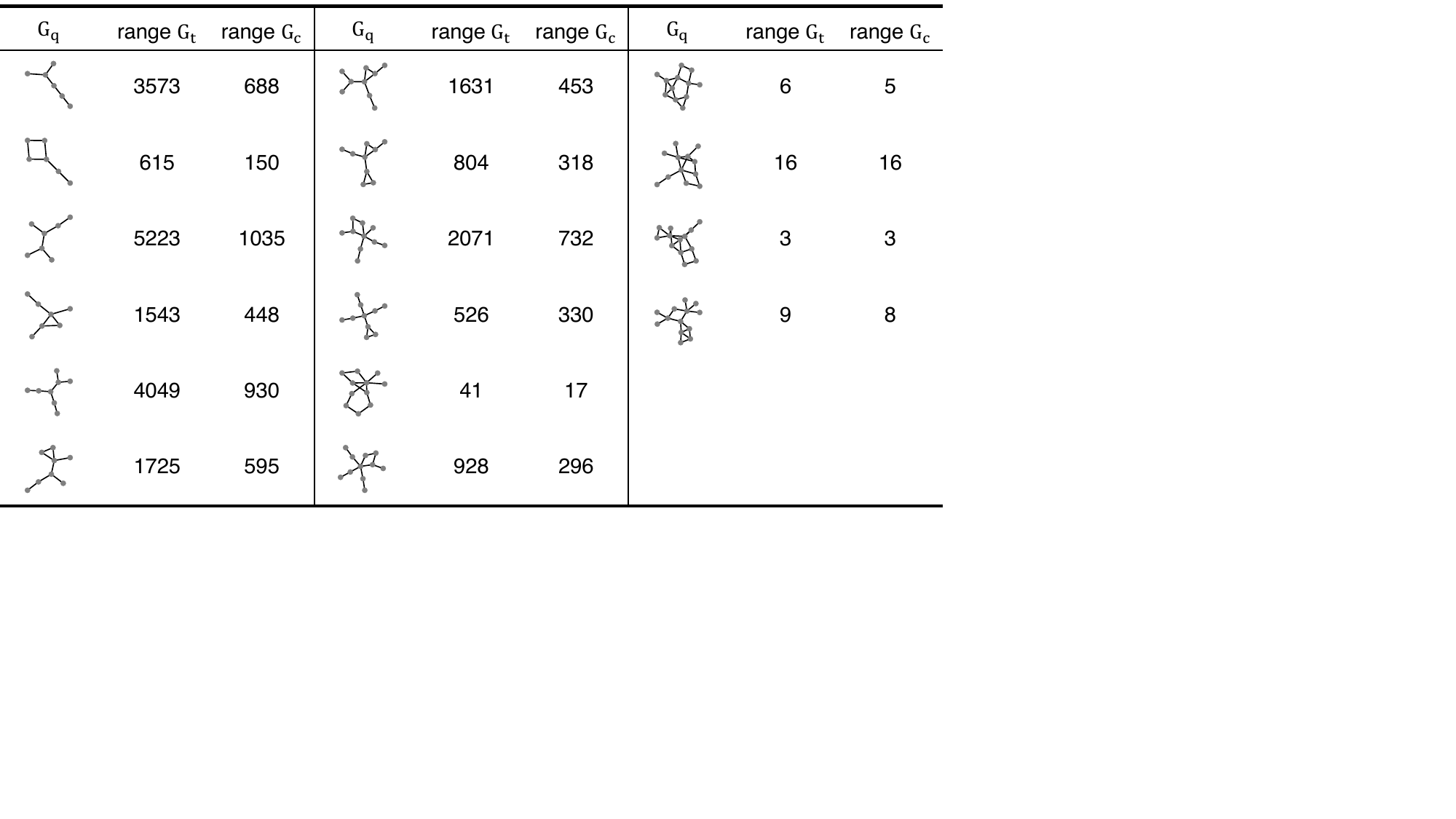}
    \vspace{-10pt}
    \caption{The \emph{large query graphs}, the range of the subgraph counts on target graphs $G_t$, and the range of the canonical counts on neighborhoods $G_c$. The statistics come from the ENZYMES dataset.}
    \label{fig:appendix/large_query_example}
\end{figure*}

Figure~\ref{fig:appendix/std_query_example} shows all twenty-nine \emph{standard queries} discussed in Section~\ref{sec:exp/setup}. They form the complete set of all non-isomorphic, connected, undirected graphs with three to five nodes. 
Figure~\ref{fig:appendix/large_query_example} shows all sixteen \emph{large query graphs} discussed in Section~\ref{sec:exp/scale}. They are frequent subgraphs with six to thirteen nodes in the ENZYMES dataset.

The figures also show the range of the ground truth counts of these queries on different target graphs from the ENZYMES dataset. The range of canonical counts on the corresponding neighborhoods is also shown. Note how \emph{canonical partition} reduces the range of counts for the regression task of GNNs.

\subsection{Target Graphs}
\label{sec:appendix/target_graphs}

To demonstrate \name's generalization power, we use real-world datasets from various domains for evaluation.
All the datasets are treated as undirected graphs without node or edge features in alignment with the setting of the approximate heuristic method~\cite{bressan2019motivo, bressan2021faster}. 
Figure~\ref{fig:syn_neigh_comprehensive} shows the graph statistics of the canonical neighborhoods from the real-world datasets and the Synthetic dataset. The Synthetic dataset successfully covers most of the canonical neighborhoods for many graph statistics, which confirms the conclusion in Figure~\ref{fig:tsne}.

\subsection{Hyper-parameter Configurations}
\label{sec:appendix/hyper_parameter}

\xhdr{\name configurations}
For \name's canonical partition stage, we set $d=4$ for all the tasks according to Theorem~\ref{thm:count_partition_count}.
For \name's neighborhood counting stage, it contains two GNNs to encode target and query graphs into embedding vectors, and a regression model to predict canonical count based on the vectors. 
For the GNN encoders, we use the triangle-based message passing variant of GraphSAGE as shown in Table~\ref{tab:appendix/shmp}. The SHMP GNN has 8 layers with a feature size of 64. The canonical node of the neighborhood is marked with a special node type. The adjacent matrix $A$ is used to find the triangle and define the heterogeneous edge type with Equation~\ref{eq:appendix/triangle}.

\begin{equation}
    E_{triangle} = \{(i,j) | (A\odot A^2)_{ij} > 0 \}
    \label{eq:appendix/triangle}
\end{equation}

For the Multilayer perceptron (MLP) of neighborhood counting, we use two fully-connected linear layers with 256 hidden feature size and LeakyReLu activation. 

For the gossip propagation stage, we use a two-layer GNN with 64 hidden feature size and a learnable gate as described in Equation~\ref{eq:GMP}. The learnable gate is a two-layer, 64-hidden-size MLP that takes the query embedding vector from the neighborhood counting stage and outputs the gate values for each GNN layer. 
The neighborhood counting prediction is expanded to 64 dimensions with a Linear layer and concatenated with the query embedding as the input for the two-layer GNN.

\xhdr{Neural baseline configurations}
We follow the configurations of the official implementations of neural baselines and adapt them to our settings. They both contain two GNN encoders and a regression model like \name's neighborhood counting model. 

For LRP, we follow the official configurations for the ZINC dataset to use a deep LRP-7-1 graph embedding layer with 8 layers and hidden dimension 8. The regression model is the same as \name.

For DIAMNet, we follow the official configurations for the MUTAG dataset to use GIN with feature size 128 as GNN encoders. The number of GNN layers is expanded from 3 to 5. The regression model is DIAMNet with 3 recurrent steps, 4 external memories, and 4 attention heads.

For DMPNN, we straightly use the official implementation of DMPNN with the official configurations for the MUTAG dataset. It use a 3-layer DMPNN.

\xhdr{Training details}
For LRP and DIMANet, We use $C_c \leftarrow \log_2(C_c+1)$ normalization for the ground truth canonical count $C_c$ to ease the high count variation problem. When evaluating the MSE of predictions, $\hat{C}_c \leftarrow 2^{\hat{C}_c}-1$ is used to undo the normalization.
We use the SmoothL1Loss with $\beta=1.0$ from PyTorch~\cite{paszke2019pytorch} as the loss function to perform the regression task of neighborhood counting and gossip propagation.

\begin{equation}
\mathcal{L}(\hat{C}_c,C_c)=
\begin{cases}
0.5(\hat{C}_c-C_c)^2 & |\hat{C}_c-C_c|<1\\
|\hat{C}_c-C_c| - 0.5 & \text{otherwise}
\end{cases}
\end{equation}

We use the Adam optimizer for neighborhood counting and gossip propagation and set the learning rate to $0.001$.

For DMPNN, we adopt the official implementation with its prescribed configurations for the MUTAG dataset. While training with the Synthetic dataset, the ground truth information of 8 queries of size 5 exceeds the maximum size that DMPNN support due to its scalability limit. Consequently, we omit these queries from the training set. This omission might marginally affect the accuracy of size 5 query predictions, but its impact on other query sizes remains minimal.

We align the computational resources when training different neural methods. \name, DIAMNet, and DMPNN have similar training efficiency, so \name's neighborhood counting model, DIAMNet and DMPNN are both trained for 300 epochs. After training the neighborhood counting model, \name's gossip propagation model is trained for 50 epochs with little resource consumption. In contrast, LRP is much slower. Even given twice training time, it can only be trained for 50 epochs. 

\xhdr{Approximate heuristic configurations}
For the MOTIVO baseline, we follow the official setting and use $10^7$ samples for each dataset. If the dataset has many graphs, the samples are evenly distributed on each target graph.

\section{Additional Numeric Results}
All of the average results are calculated by geometric mean.

\label{sec:appendix/numeric_result}

\subsection{Ablation study}
We perform an ablation study across 8 real-world datasets to demonstrate the effectiveness of our canonical partition, SHMP, and gossip propagation components. Numerical results are presented in Table~\ref{tab:appendix/ablation_canonical_partition}, Table~\ref{tab:appendix/shmp}, and Table~\ref{tab:appendix/gossip}. The ablation study shows that all three components are essential for \name's performance.

\begin{table*}[tb]
    \centering
    \resizebox{\linewidth}{!}{
    \begin{tabular}{lccc|ccc|ccc|ccc}
        \toprule
        Dataset & \multicolumn{3}{c|}{MUTAG} & \multicolumn{3}{c|}{COX2} & \multicolumn{3}{c|}{ENZYMES} & \multicolumn{3}{c}{IMDB-BINARY}\\
        Query-Size & 3 & 4 & 5 & 3 & 4 & 5 & 3 & 4 & 5 & 3 & 4 & 5 \\ 
        \midrule
        \multicolumn{13}{c}{normalized MSE} \\    
        \midrule
        w/o $\mathcal{P}$ & 6.0E-1 & 3.1E-1 & 3.4E-1 & 1.3E+0 & 6.8E-1 & 6.8E-1 & 5.3E-1 & 3.7E-1 & 4.6E-1 & 9.5E+07 & 1.0E+10 & 3.5E+14 \\
        w $\mathcal{P}$ & \textbf{1.8E-2} & \textbf{1.6E-2} & \textbf{1.7E-2} & \textbf{1.1E-2} &\textbf{ 1.2E-2} & \textbf{9.3E-3} & \textbf{1.1E-2} & \textbf{3.7E-2} & \textbf{4.7E-2} & \textbf{7.9E-03} & \textbf{2.1E-01} & \textbf{4.5E-01} \\
        \midrule
        \multicolumn{13}{c}{MAE} \\
        \midrule
        w/o $\mathcal{P}$ & 8.3E+0 & 4.8E+0 & 3.6E+00 & 3.0E+1 & 1.6E+1 & 9.9E+0 & 3.4E+1 & 3.5E+1 & 3.2E+1 & 2.3E+05 & 1.4E+07 & 1.8E+10\\
        w $\mathcal{P}$ & \textbf{1.4E-1} & \textbf{9.8E-1} & \textbf{4.8E-1} & \textbf{2.7E+0} & \textbf{2.3E+0} & \textbf{1.2E+0} & \textbf{5.5E+0} & \textbf{9.3E+0} & \textbf{9.5E+0} & \textbf{2.5E+01} & \textbf{3.0E+02} & \textbf{1.6E+03}\\
        \midrule
        \midrule
        Dataset & \multicolumn{3}{c|}{MSRC-21} & \multicolumn{3}{c|}{CiteSeer} & \multicolumn{3}{c|}{Cora} & \multicolumn{3}{c}{FIRSTMM-DB} \\
        Query-Size & 3 & 4 & 5 & 3 & 4 & 5 & 3 & 4 & 5 & 3 & 4 & 5 \\ 
        \midrule
        \multicolumn{13}{c}{normalized MSE} \\
        \midrule
        w/o $\mathcal{P}$ & 8.4E+00 & 1.3E+02 & 6.7E+02 & inf & inf & inf & 3.3E+29 & 5.3E+39 & 1.0E+48 & inf & inf & inf \\
        w $\mathcal{P}$ & \textbf{7.5E-03} & \textbf{6.9E-03} & \textbf{6.7E-02} & \textbf{6.9E-05} & \textbf{1.1E-01} & \textbf{1.7E-01} & \textbf{5.3E-03} & \textbf{2.2E-01} & \textbf{7.0E-02} & \textbf{1.7E-03} & \textbf{2.3E-02} & \textbf{5.1E-02} \\
        \midrule
        \multicolumn{13}{c}{MAE} \\    
        \midrule
        w/o $\mathcal{P}$ & 4.7E+02 & 2.3E+03 & 4.3E+03 & inf & inf & inf & 9.4E+18 & 1.1E+25 & 1.6E+30 & inf & inf & inf \\
        w $\mathcal{P}$ & \textbf{1.8E+01} & \textbf{3.3E+01} & \textbf{1.2E+02} & \textbf{7.3E+01} & \textbf{1.3E+04} & \textbf{1.1E+05} & \textbf{1.4E+03} & \textbf{7.5E+04} & \textbf{5.6E+05} & \textbf{1.8E+02} & \textbf{6.7E+02} & \textbf{1.0E+03} \\
        \bottomrule
    \end{tabular}
    }
\caption{Normalized MSE performance with or without canonical partition. Since gossip propagation relies on the output of neighborhoods, it's also removed for both for a fair comparison.}
\label{tab:appendix/ablation_canonical_partition}
\vspace{-10pt}
\end{table*}

\begin{table*}[htb]
    \centering
    \resizebox{\linewidth}{!}{
    
    \begin{tabular}{lccc|ccc|ccc|ccc}
    \toprule
        Dataset & \multicolumn{3}{c|}{MUTAG} & \multicolumn{3}{c|}{COX2} & \multicolumn{3}{c|}{ENZYMES} & \multicolumn{3}{c}{IMDB-BINARY}\\
        Query-Size & 3 & 4 & 5 & 3 & 4 & 5 & 3 & 4 & 5 & 3 & 4 & 5 \\ 
        \midrule
        \multicolumn{13}{c}{normalized MSE} \\    
        \midrule
        GCN & inf & inf & inf & inf & inf & inf & inf & inf & inf & inf & inf & inf \\ 
        SAGE & 4.0E-01 & 9.6E-02 & 5.6E-01 & 3.1E-01 & 5.4E-02 & 3.8E-01 & 5.1E-01 & 3.2E-01 & 3.0E-01 & 2.9E+01 & 1.5E+01 & 4.6E+00 \\
        GIN & 8.4E-02 & 7.3E-02 & 1.6E-01 & 3.6E-02 & 6.6E-02 & 2.3E-01 & 1.9E-01 & 7.1E-02 & 1.1E-01 & 1.1E+00 & 1.0E+00 & 1.0E+00 \\
        ID-GNN & 3.3E-02 & 2.9E-02 & \textbf{1.5E-02} & 1.5E-02 & \textbf{7.0E-03} & 2.9E-02 & 1.9E-02 & \textbf{2.6E-02} & 8.1E-02 & 1.1E+00 & 1.0E+00 & 1.0E+00 \\
        \midrule
        SAGE+SHMP & \textbf{1.8E-02} & \textbf{1.6E-02} & 1.7E-02 & \textbf{1.1E-02} & 1.2E-02 & \textbf{1.0E-02} & \textbf{1.1E-02} & 3.7E-02 & \textbf{4.7E-02} & \textbf{7.9E-03} & \textbf{2.1E-01} & \textbf{4.5E-01} \\
        \midrule
        \multicolumn{13}{c}{MAE} \\
        \midrule
        GCN & inf & inf & inf & inf & inf & inf & inf & inf & inf & inf & inf & inf \\ 
        SAGE & 6.8E+00 & 2.3E+00 & 7.4E+00 & 1.4E+01 & 4.3E+00 & 1.8E+01 & 4.1E+01 & 3.4E+01 & 2.9E+01 & 5.9E+02 & 2.4E+03 & 6.1E+03 \\
        GIN & 3.5E+00 & 2.3E+00 & 1.7E+00 & 5.8E+00 & 6.7E+00 & 6.4E+00 & 2.5E+01 & 1.8E+01 & 1.8E+01 & 3.0E+02 & 8.3E+02 & 2.6E+03 \\
        ID-GNN & 1.9E+00 & 1.0E+00 & 5.3E-01 & 3.2E+00 & \textbf{1.7E+00} & 1.8E+00 & 6.2E+00 & \textbf{8.9E+00} & 1.3E+01 & 3.0E+02 & 8.3E+02 & 2.6E+03 \\
        \midrule
        SAGE+SHMP & \textbf{1.4E+00} & \textbf{9.8E-01} & \textbf{4.8E-01} & \textbf{2.7E+00} & 2.3E+00 & \textbf{1.2E+00} & \textbf{5.5E+00} & 9.3E+00 & \textbf{9.5E+00} & \textbf{2.5E+01} & \textbf{3.0E+02} & \textbf{1.6E+03} \\
        \midrule
        \midrule
        Dataset & \multicolumn{3}{c|}{MSRC-21} & \multicolumn{3}{c|}{CiteSeer} & \multicolumn{3}{c|}{Cora} & \multicolumn{3}{c}{FIRSTMM-DB} \\
        Query-Size & 3 & 4 & 5 & 3 & 4 & 5 & 3 & 4 & 5 & 3 & 4 & 5 \\ 
        \midrule
        \multicolumn{13}{c}{normalized MSE} \\    
        \midrule
        GCN & inf & inf & inf & inf & inf & inf & inf & inf & inf & inf & inf & inf \\ 
        SAGE & 1.9E-01 & 6.3E-01 & 3.4E-01 & 1.5E+00 & 2.7E-01 & 9.7E-01 & 9.4E+00 & 5.9E-01 & 1.1E+00 & 3.2E-02 & 1.7E-01 & 2.2E-01 \\
        GIN & 2.4E+00 & 1.6E+00 & 1.3E+00 & 1.9E+00 & 1.5E+00 & 1.2E+00 & 1.9E+00 & 1.3E+00 & 1.1E+00 & 1.1E+00 & 1.2E+00 & 1.1E+00 \\
        ID-GNN & 3.0E+00 & 1.7E+00 & 1.3E+00 & 2.2E+00 & 1.5E+00 & 1.2E+00 & 2.1E+00 & 1.3E+00 & 1.1E+00 & 1.5E+00 & 1.2E+00 & 1.1E+00 \\
        \midrule
        SAGE+SHMP & \textbf{7.5E-03} & \textbf{6.9E-03} & \textbf{6.7E-02} & \textbf{6.9E-05} & \textbf{1.1E-01} & \textbf{1.7E-01} & \textbf{5.3E-03} & \textbf{2.2E-01} & \textbf{7.0E-02} & \textbf{1.7E-03} & \textbf{2.3E-02} & \textbf{5.1E-02} \\
        \midrule
        \multicolumn{13}{c}{MAE} \\
        \midrule
        GCN & inf & inf & inf & inf & inf & inf & inf & inf & inf & inf & inf & inf \\ 
        SAGE & 9.6E+01 & 3.0E+02 & 3.7E+02 & 9.7E+03 & 2.7E+04 & 3.1E+05 & 5.5E+04 & 1.7E+05 & 2.1E+06 & 8.3E+02 & 2.3E+03 & 2.6E+03 \\
        GIN & 3.1E+02 & 4.9E+02 & 6.4E+02 & 1.1E+04 & 6.0E+04 & 3.7E+05 & 2.3E+04 & 2.2E+05 & 2.2E+06 & 3.8E+03 & 4.8E+03 & 4.9E+03 \\
        ID-GNN & 3.6E+02 & 5.0E+02 & 6.4E+02 & 1.2E+04 & 6.0E+04 & 3.7E+05 & 2.5E+04 & 2.2E+05 & 2.2E+06 & 4.6E+03 & 5.0E+03 & 4.9E+03 \\
        \midrule
        SAGE+SHMP & \textbf{1.8E+01} & \textbf{3.3E+01} & \textbf{1.2E+02} & \textbf{7.3E+01} & \textbf{1.3E+04} & \textbf{1.1E+05} & \textbf{1.4E+03} & \textbf{7.5E+04} & \textbf{5.6E+05} & \textbf{1.8E+02} & \textbf{6.7E+02} & \textbf{1.0E+03} \\
    \bottomrule
    \end{tabular}
    }
    \caption{Normalized MSE and MAE performance with different GNN models for neighborhood counting.}
    \label{tab:appendix/shmp}
    \vspace{-10pt}
\end{table*}

\begin{table*}[tb]
    \centering
    \resizebox{\linewidth}{!}{
    \setlength\tabcolsep{4pt} 
    \begin{tabular}{lccc|ccc|ccc|ccc}
        \toprule
        Dataset & \multicolumn{3}{c|}{MUTAG} & \multicolumn{3}{c|}{COX2} & \multicolumn{3}{c|}{ENZYMES} & \multicolumn{3}{c}{IMDB-BINARY}\\
        Query-Size & 3 & 4 & 5 & 3 & 4 & 5 & 3 & 4 & 5 & 3 & 4 & 5 \\ 
        \midrule
        \multicolumn{13}{c}{normalized MSE} \\    
        \midrule
        w/o propagation & 1.8E-02 & 1.6E-02 & 1.7E-02 & 1.1E-02 & 1.2E-02 & 1.0E-02 & 1.1E-02 & \textbf{3.7E-02} & \textbf{4.7E-02} & \textbf{7.9E-03} & \textbf{2.1E-01} & \textbf{4.5E-01} \\
        w propagation & \textbf{2.3E-03} & \textbf{8.4E-04} & \textbf{6.5E-03} & \textbf{6.9E-04} & \textbf{5.3E-04} & \textbf{5.4E-03} & \textbf{5.3E-03} & 5.7E-02 & 5.3E-02 & 8.7E-03 & 2.1E-01 & 4.5E-01 \\
        \midrule
        \multicolumn{13}{c}{MAE} \\
        \midrule
        w/o propagation & 1.4E+00 & 9.8E-01 & 4.8E-01 & 2.7E+00 & 2.3E+00 & 1.2E+00 & 5.5E+00 & \textbf{9.3E+00} & \textbf{9.5E+00} & 2.5E+01 & \textbf{3.0E+02} & \textbf{1.6E+03} \\
        w propagation & \textbf{5.1E-01} & \textbf{1.9E-01} & \textbf{3.0E-01} & \textbf{6.2E-01} & \textbf{4.0E-01} & \textbf{8.1E-01} & \textbf{3.5E+00} & 1.1E+01 & 9.8E+00 & \textbf{2.3E+01} & 3.0E+02 & 1.6E+03 \\
        \midrule
        \midrule
        Dataset & \multicolumn{3}{c|}{MSRC-21} & \multicolumn{3}{c|}{CiteSeer} & \multicolumn{3}{c|}{Cora} & \multicolumn{3}{c}{FIRSTMM-DB}\\
        Query-Size & 3 & 4 & 5 & 3 & 4 & 5 & 3 & 4 & 5 & 3 & 4 & 5 \\ 
        \midrule
        \multicolumn{13}{c}{normalized MSE} \\
        \midrule
        w/o propagation & 7.5E-03 & 6.9E-03 & \textbf{6.7E-02} & 6.9E-05 & 1.1E-01 & 1.7E-01 & 5.3E-03 & 2.2E-01 & 7.0E-02 & \textbf{1.7E-03} & \textbf{2.3E-02} & \textbf{5.1E-02} \\
        w propagation & \textbf{2.6E-03} & \textbf{3.9E-03} & 8.5E-02 & \textbf{3.5E-05} & \textbf{9.7E-02} & \textbf{1.6E-01} & \textbf{4.2E-03} & \textbf{2.1E-01} & \textbf{6.3E-02} & 2.1E-03 & 3.4E-02 & 5.4E-02 \\
        \midrule
        \multicolumn{13}{c}{MAE} \\    
        \midrule
        w/o propagation & 1.8E+01 & 3.3E+01 & \textbf{1.2E+02} & 7.3E+01 & 1.3E+04 & 1.1E+05 & 1.4E+03 & 7.5E+04 & 5.6E+05 & 1.8E+02 & \textbf{6.7E+02} & 1.0E+03 \\
        w propagation & \textbf{1.0E+01} & \textbf{2.5E+01} & 1.3E+02 & \textbf{6.0E+01} & \textbf{1.2E+04} & \textbf{1.1E+05} & \textbf{1.3E+03} & \textbf{7.3E+04} & \textbf{5.4E+05} & \textbf{1.5E+02} & 7.2E+02 & \textbf{9.6E+02} \\
        \bottomrule
    \end{tabular}
    }
    \caption{The normalized MSE and MAE performance with and without gossip propagation.}
    \label{tab:appendix/gossip}
    \vspace{-20pt}
\end{table*}

\begin{table*}[tb]
    \centering
    \begin{tabular}{lccc|ccc|ccc}
    \toprule
        Dataset & \multicolumn{3}{c|}{CiteSeer} & \multicolumn{3}{c|}{Cora} & \multicolumn{3}{c}{FIRSTMM-DB}\\ 
        Query-Size & 3 & 4 & 5 & 3 & 4 & 5 & 3 & 4 & 5 \\ \midrule
        \multicolumn{10}{c}{normalized MSE} \\
        \midrule
        LRP & inf & inf & inf & inf & inf & inf & 1.5E+00 & 1.2E+00 & 1.1E+00 \\ 
        DIAMNet & 2.0E+00 & 1.5E+00 & 1.2E+00 & 1.0E+10 & 3.2E+07 & 3.7E+04 & 1.7E+14 & 5.5E+13 & 9.7E+12 \\
        DMPNN & 9.5E+04 & 2.5E+02 & 6.8E+01 & 1.8E+05 & 1.1E+02 & 6.7E+01 & 5.6E+02 & 2.5E+02 & 1.6E+02 \\
        \midrule
        \name & \textbf{3.5E-05} & \textbf{9.7E-02} & \textbf{1.6E-01} & \textbf{4.2E-03} & \textbf{2.1E-01} & \textbf{6.3E-02} & \textbf{2.1E-03} & \textbf{3.6E-02} & \textbf{5.4E-02} \\
        \midrule
        \multicolumn{10}{c}{MAE} \\
        \midrule
        LRP & inf & inf & inf & inf & inf & inf & 4.5E+03 & 4.9E+03 & 4.9E+03 \\ 
        DIAMNet & 1.1E+04 & 6.0E+04 & 3.6E+05 & 2.1E+09 & 1.6E+09 & 8.3E+08 & 1.7E+10 & 1.3E+10 & 6.6E+09 \\
        DMPNN & 6.1E+06 & 7.6E+06 & 8.7E+06 & 1.8E+07 & 2.4E+07 & 3.0E+07 & 2.2E+05 & 2.9E+05 & 3.3E+05 \\
        \midrule
        \name & \textbf{6.0E+01} & \textbf{1.2E+04} & \textbf{1.1E+05} & \textbf{1.3E+03} & \textbf{7.3E+04} & \textbf{5.4E+05} & \textbf{1.5E+02} & \textbf{7.3E+02} & \textbf{9.6E+02} \\
    \bottomrule
    \end{tabular}
    \caption{Normalized MSE and MAE performance of neural methods on large targets with standard queries.}
    \label{appendix/tab:larget}
    \vspace{-8pt}
\end{table*}

\subsection{Count Distribution Prediction} 
\label{sec:appendix/distribution}

To the best of our knowledge, \name is the first approximate method that predicts the subgraph count distribution over the whole target graph. 
We use the canonical count of each node as the ground truth for the distribution prediction accuracy analysis. 
The canonical count represents the number of \emph{pattern}s in each node's neighborhood while avoiding missing or double counting as discussed in Section~\ref{sec:pipeline/canonical_count}.
Following the setup in Section~\ref{sec:exp/setup}, we use all the size $3-5$ \emph{standard query} graphs to test the distribution performance of \name on different target graphs. 
The normalized MSE is the mean square error of the canonical count prediction of each (query, target graph node) pair divided by the variance of the (query, target graph node) pair's true canonical count.
The MAE is the mean absolute error of the canonical count prediction of each (query, target graph node) pair.

\begin{table*}[!ht]
    \centering
    \begin{tabular}{lccc|ccc|ccc}
    \toprule
        Dataset & \multicolumn{3}{c|}{MUTAG} & \multicolumn{3}{c|}{COX2} & \multicolumn{3}{c}{ENZYMES} \\
        Query-Size & 3 & 4 & 5 & 3 & 4 & 5 & 3 & 4 & 5 \\
        \midrule    
        Norm. MSE & 7.51E-2 & 2.36E-1 & 1.71E+0 & 4.94E-4 & 5.63E-4 & 1.44E-2 & 4.74E-5 & 5.69E-5 & 5.66E-4 \\
        MAE & 2.97E-4 & 1.19E-3 & 1.90E-2 & 3.90E-4 & 5.19E-4 & 1.71E-2 & 7.60E-2 & 1.51E-1 & 3.15E-1 \\ 
    \bottomrule
    \end{tabular}
    \caption{\name's count distribution prediction error under normalized MSE and MAE. Use the canonical count of each target graph node as the ground truth.}
    \label{tab:appendix/distribution}
\end{table*}

Experiments show \name achieves a low $3.8 \times 10^{-3}$ normalized MSE for the count distribution prediction task. See Table~\ref{tab:appendix/distribution} for the detailed results.
A visualization of \name's distribution prediction on the CiteSeer dataset is also shown in Figure~\ref{fig:distribution}.
Note how \name accurately predicts the distribution while providing meaningful insight on the graph.

\section{Runtime Comparision} 
\label{sec:appendix/runtime}

\subsection{Experiment setup}
\label{sec:appendix/runtime/setup}

\begin{figure}[tb]%
    \centering
    \includegraphics[width=0.35\textwidth]{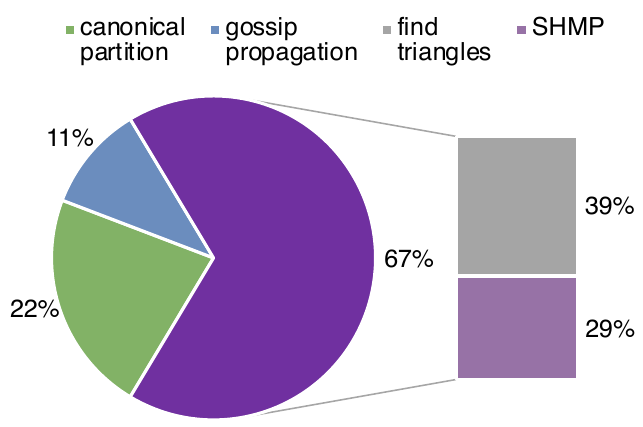}
    \vspace{-7pt}
    \caption{Runtime breakdown of \name.}
    \label{fig:runtime_breakdown}
    \vspace{-8pt}
\end{figure}

We configure \name and baselines as follows.
For the exact method VF2~\cite{cordella2004sub}, we use the Python implementation from the graph processing framework~\cite{hagberg2008networkx} and use Python's concurrent standard library to enable multiprocessing on four CPU cores.
For the exact method IMSM~\cite{sun2020memory}, we use the official c++ implementation with four CPU cores. We use the IMSM-recommended method configurations: GQL~\cite{he2008GQL} as the filtering method, RI~\cite{bonnici2013RI} as the ordering method, and LFTJ~\cite{bhattarai2019ceci, han2019efficient} as the enumeration method. The failing set pruning optimization is also enabled.
For the approximate heuristic method MOTIVO~\cite{bressan2019motivo}, we use the optimized c++ implementation from~\cite{bressan2021faster} with four CPU cores.
For the neural method \name, we use the Python implementation with one CPU core and one GPU core.

We use Intel Xeon Gold 6226R CPU with 2.90GHz frequency and NVIDIA GeForce RTX 3090 GPU for runtime tests. All the methods are set to count the induced subgraphs in the ENZYMES dataset. 
Note that IMSM can only perform non-induced subgraph counting.
So VF2, MOTIVO, and \name are set to perform induced subgraph counting tasks, while IMSM performs non-induced tasks for runtime comparison.
For query sizes no larger than five nodes, the \emph{standard queries} from Section~\ref{sec:exp/setup} are used. For query sizes larger than five, the same thirty queries of each size are selected for VF2 and \name. We cannot assign specific queries for MOTIVO, so it is set to output the count of any thirty queries of each size.
In the experiments, the data loading and graph format conversion time is ignored for all methods. We further extend the time budget for MOTIVO to 60 minutes. The results show that \name achieves 15$\times$, 53$\times$, and 120$\times$ speedup over MOTIVO for query size 13 to 15, respectively. 

As Figure~\ref{fig:runtime_breakdown} shows, currently \name's triangle finding in \emph{neighborhood counting} takes the majority of the runtime, which can be easily substituted with other efficient implementations, e.g.,~\cite{Donato2018TriangleCW}, to further speed up \name.

\subsection{Asymptotic complexity}
\label{sec:appendix/asymptotic_complexity}

For the proposed \name's three-step pipeline, assuming the average canonical neighborhood $G_n$ of the target graph $G_t$ has $V_n$ nodes and $E_n$ edges. The time complexity for canonical partition is the index-restricted breadth-first search starting from all the target vertices as shown in Appendix~\ref{sec:appendix/implement_canonical_partition}, which is $O(V_t\times (\bar{V}_n+\bar{E}_n))$. 
The time complexity for neighborhood counting consists of triangle counting and heterogeneous message passing on $G_q$ and $G_t$. 
The complexity of triangle counting is $O(E^{3/2})$ on the target and query graph~\cite{itai1977finding}.
The heterogeneous message passing has the complexity of regular GNNs~\cite{maron2019provably} on the $V_t$ neighborhoods and the queries, which is $O(V_t\times (\bar{V}_n+\bar{E}_n)) + O(V_q+E_q)$. 
For gossip propagation, the time complexity also equals a regular GNN, which is $O(E_t+V_t)$. 

In conclusion, the overall time complexity of \name is $O(E_t^{3/2}+V_t\times (\bar{V}_n+\bar{E}_n)) + O(E_q^{3/2}+V_q)$. In real-world graphs, the common contraction of neighborhoods~\cite{Weber2019CurvatureAR} makes $\bar{V}_n$ and $\bar{E}_n$ relatively small. So the major asymptotic complexity comes from the triangle counting on the target graph, which only has polynomial time complexity.

In contrast, for the heuristic approximate method MOTIVO, the build-up phase alone has time complexity $O(a^{V_q}\times E_t)$ for some $a>0$. So it suffers from exponential runtime growth.
For exact method VF2, the time complexity is $O(V^2)$ to $O(V!\times V)$ where $V=\max\{V_t,V_q\}$. In practice, we generally observe exponential runtime growth. Experiments of Figure~\ref{fig:exp/runtime} confirm the above analysis.

\subsection{Discussion on GPU-based exact methods}
\label{sec:appendix/gpu_exact}
Like CPU-based exact methods, GPU-based exact methods also suffer from high asymptotic complexity, thus not scalable.
We wish to provide a precise comparison with existing GPU-based exact methods. Unfortunately, existing method~\cite{lin2016network} does not open-source their code.
Thus, we directly compare the reported results and find that our method can easily beat it. For example, when both are set to find size-4 to size-8 queries on the $10^4$-scale target graphs (CE and ENZYMES), ~\cite{lin2016network} takes $1\times 10^4$ seconds (Figure 4(a) in~\cite{lin2016network}), while \name only takes $1.7\times 10^2$ seconds. Though the query graphs are not exactly the same, \name is much faster in general.

\begin{table*}[htb]
    \centering
    \resizebox{\linewidth}{!}{
    \setlength\tabcolsep{3pt}
    \begin{tabular}{lccc|ccc|ccc|ccc|ccc|ccc}
        \toprule
        Test Dataset & \multicolumn{3}{c|}{IMDB-BINARY} & \multicolumn{3}{c|}{MSRC-21} & \multicolumn{3}{c|}{CiteSeer} & \multicolumn{3}{c|}{Cora} & \multicolumn{3}{c|}{FIRSTMM-DB} & \multicolumn{3}{c}{MUTAG}\\ 
        Query-Size & 3 & 4 & 5 & 3 & 4 & 5 & 3 & 4 & 5 & 3 & 4 & 5 & 3 & 4 & 5 & 3 & 4 & 5 \\ 
        \midrule
        \multicolumn{19}{c}{normalized MSE} \\
        \midrule
        Pre-train on Existing & \multicolumn{3}{c|}{overflow} & 1.1E+01 & 1.9E+00 & 1.1E+00 & 1.7E-01 & 5.1E-01 & 6.8E-01 & 4.4E-01 & 1.1E+00 & 8.0E-01 & 1.1E-01 & 1.1E-01 & 1.6E-01 & 6.51E-03 & 3.40E-03 & 8.70E-02 \\
        Pre-train on Synthetic & 8.5E-03 & 2.1E-01 & 4.5E-01 & 2.5E-03 & 3.8E-03 & 8.7E-02 & 3.5E-05 & 9.7E-02 & 1.6E-01 & 4.2E-03 & 2.1E-01 & 6.3E-02 & 2.1E-03 & 3.6E-02 & 5.4E-02 & 2.3E-03 & 8.4E-04 & 6.5E-03 \\
        \midrule
        \multicolumn{19}{c}{MAE} \\
        \midrule
        Pre-train on Existing & \multicolumn{3}{c|}{overflow} & 2.9E+02 & 4.5E+02 & 6.1E+02 & 3.3E+03 & 3.4E+04 & 2.5E+05 & 1.1E+04 & 1.7E+05 & 1.6E+06 & 1.2E+03 & 1.7E+03 & 2.2E+03 & 7.85E-01 & 3.68E-01 & 1.40E+00 \\
        Pre-train on Synthetic & 2.4E+01 & 3.0E+02 & 1.6E+03 & 1.0E+01 & 2.5E+01 & 1.3E+02 & 6.0E+01 & 1.2E+04 & 1.1E+05 & 1.3E+03 & 7.3E+04 & 5.4E+05 & 1.5E+02 & 7.3E+02 & 9.6E+02 & 5.1E-01 & 1.9E-01 & 3.0E-01 \\
        \bottomrule
    \end{tabular}%
    }
    \caption{Normalized MSE and MAE performance with different training datasets. When pre-training on existing datasets, IMDB-BINARY and MSRC-21 use MUTAG; CiteSeer uses Cora; Cora and FIRSTMM-DB use CiteSeer; MUTAG uses ENZYMES.}
    \label{tab:app/general_small}
    \vspace{-10pt}
\end{table*}

\section{Additional Results Analysis for Large Queries}
\label{sec:appendix/other_metric}

To give an even more in-depth understanding of the performance for large queries, we additionally provide the results with more evaluation metrics.

\subsection{Q-error Analysis} \label{sec:appendix/q_error}

\xhdr{Definition}
Given the ground truth subgraph count $\mathcal{C}$ of query $G_q$ in target $G_t$, as well as the estimated count $\hat{\mathcal{C}}$. We use the definition of q-error from previous work~\cite{Zhao2021ALS}.

\begin{equation}
    e_q(G_q, G_t) = \max\left\{ \frac{\mathcal{C}}{\hat{\mathcal{C}}}, \frac{\hat{\mathcal{C}}}{\mathcal{C}} \right\}, 
    \space e_q \in [1,+\inf)
    \label{eq:appendix/absolute-q-error}
\end{equation}

The q-error quantifies the factor that the estimation differs from the true count. The more it is close to $1$, the better estimation.
In~\cite{Zhao2021ALS}, there is also an alternative form of q-error used in figures to show the systematic bias of predictions.

\begin{equation}
    e_q(G_q, G_t) =  \frac{\hat{\mathcal{C}}}{\mathcal{C}},
    \space e_q \in (0,+\inf)
    \label{eq:appendix/q-error}
\end{equation}

We follow the previous settings and use Equation~\ref{eq:appendix/q-error} in our visualization.

\begin{figure}[h]
    \centering
    \includegraphics[width=0.5\textwidth]
    {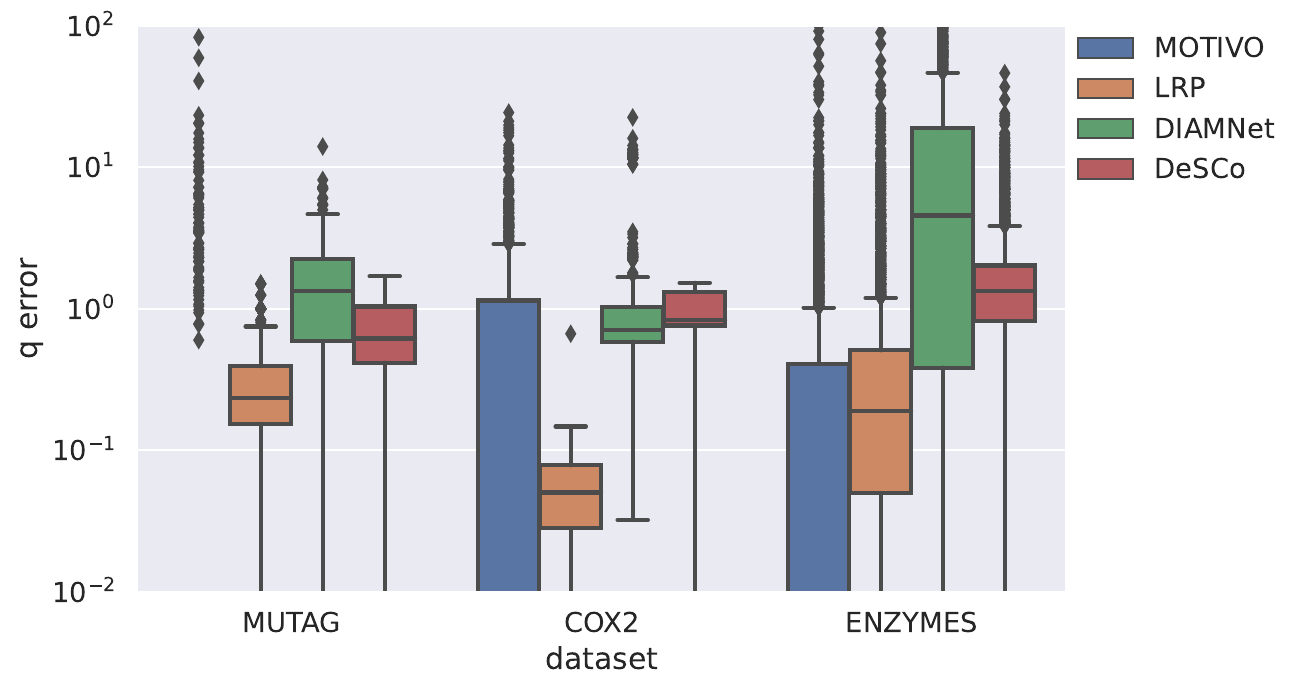}
    \caption{The q-error box plot of large query-target pairs. The q-error (y-axis) is clipped at $10^{-2}$ and $10^2$. For q-error, the closer to 1, the better.}
    \label{fig:appendix/large_query_q_error_box}
\end{figure}

\xhdr{Experimental results}
We reassess the performance of \name with large queries. The results are shown in Figure~\ref{fig:appendix/large_query_q_error_box}. 
The data that $\mathcal{C}=0$ is ignored for mathematic correctness. 
The box of MOTIVO on MUTAG is too close to zero to be shown in the figure.
DeSCo's q-error is the closest to 1 with minimal spread. It shows how \name excels in systematic error and consistency compared with the baselines.

\begin{table}[htb]
    \centering
    \begin{tabular}{lc|c|c}
        \toprule
        Dataset & MUTAG & COX2 & ENZYMES \\
        \midrule
        MOTIVO & 1.2E+01 & 3.2E+00 & 1.4E+00 \\ 
        LRP & 7.6E-01 & 1.1E+00 & 6.7E-01 \\ 
        DIAMNet & 2.3E+00 & 2.4E-01 & 8.9E+01 \\ 
        \midrule
        \name & \textbf{1.5E-01} & \textbf{1.2E-01} & \textbf{4.0E-01} \\ 
        \bottomrule
    \end{tabular}
    \caption{Normalized MSE of approximate heuristic and neural methods on subgraph counting of sixteen large queries.}
    \label{tab:appendix/large_query_mse}
\end{table}

\xhdr{Limitations of q-error}
Despite the advantage of demonstrating relative error, the q-error metric also has obvious limitations, thus not being chosen as the major evaluation metric. 
In~\cite{Zhao2021ALS}, the authors assume $\mathcal{C} \geq 1$ and $\hat{\mathcal{C}} \geq 1$. However, this assumption may not hold, given that the query graph may not (or is predicted to) exist in the target graph, especially for larger queries. 
The zero or near-zero denominators greatly influence the average q-error. It causes the overestimation of the subgraph existence problem instead of the subgraph counting problem.

\subsection{MSE Analysis}
\label{sec:appendix/mse}

\xhdr{Definition}
We follow the same setting in Figure~\ref{fig:exp/large_query} and show the normalized MSE for predicting the subgraph count of large queries. Note that in a few cases, the tested large queries of a certain size may not exist in the target graph. For example, the two size-thirteen queries in Figure~\ref{fig:appendix/large_query_example} do not exist in the CiteSeer dataset. To prevent divide-by-zero in normalization, the MSE is normalized by the variance of ground truth counts of all large queries, instead of being normalized for each query size.

\xhdr{Experimental results} The experimental results are shown in Table~\ref{tab:appendix/large_query_mse}. \name demonstrates the lowest MSE on all tested target graphs.

\subsection{Observation of Prediction Error}
Based on the experimental results across different queries, we observe a positive correlation between MAE and the queriy size. 
Furthermore, for queries with a larger ground truth count, the MAE tends to increase.
Among queries of equivalent size, those with fewer edges exhibit a higher error. This phenomenon could be attributed to the fact that such queries often have a larger ground truth count, indicating a potential underlying complexity in accurate prediction.

\section{Future Work}

While \name significantly advances neural methods in processing large target graphs, it's important to note that certain heuristic methods remain more efficient for counting target graphs with millions of nodes, particularly for smaller queries. The ability of \name to efficiently handle exceptionally large target graphs is a significant development. However, evaluating its effectiveness on million-node scale graphs is challenging, mainly due to the substantial overhead in obtaining accurate ground truth counts with exact counting methods. Therefore, developing methods for efficiently evaluating approximate subgraph counting on such a grand scale is an essential area for future research.

Additionally, \name shows remarkable adaptability for unseen target graphs. However, extending this adaptability to unseen queries, especially in zero-shot settings, poses an intriguing challenge and is a promising area for further study. Furthermore, the potential of neural methods to handle larger queries with increased diameters remains an exciting avenue for future exploration.

Moreover, \name's canonical partition scheme, which significantly enhances the accuracy of neural subgraph counting, might also be beneficial for exact and heuristic counting methods. This scheme's efficacy is partly linked to the indexing strategy of nodes, with \name currently utilizing random indexing to ensure robust performance. Investigating alternative indexing strategies could offer valuable insights and improvements.

By exploring these areas, we can potentially expand the applications of scalable and adaptable subgraph counting, pushing the boundaries of current methodologies.

\end{document}